\documentclass[letterpaper, 10 pt, journal, twoside]{IEEEtran}
\usepackage{amsmath,amssymb} 
\usepackage{graphicx} 
\usepackage{cite} 
\usepackage{url} 
\usepackage{booktabs} 
\usepackage{tabularx} 
\newcolumntype{Y}{>{\centering\arraybackslash}X} 
\usepackage{multirow} 
\usepackage{xcolor} 
\usepackage{bm} 
\usepackage{xspace} 
\usepackage{tikz} 
\usepackage{pgfplots} 
\pgfplotsset{compat=1.17} 
\usetikzlibrary{arrows.meta, positioning, calc} 
\usepgfplotslibrary{groupplots,fillbetween}
\usepackage{orcidlink}
\usepackage{hyperref}
\definecolor{okBlue}{HTML}{0072B2}
\definecolor{okOrange}{HTML}{E69F00}
\definecolor{okGreen}{HTML}{009E73}
\definecolor{okRed}{HTML}{D55E00}
\definecolor{okPurple}{HTML}{CC79A7}
\definecolor{okSky}{HTML}{56B4E9}
\definecolor{okYellow}{HTML}{F0E442}
\definecolor{okGrey}{HTML}{7F7F7F}
\ifdefined\pgfplotsset
\pgfplotsset{
  m hiper/.style={okBlue, line width=1.3pt, mark=*, mark size=1.4pt},
  m pf/.style={okRed, line width=0.7pt, mark=triangle*, mark size=1.6pt},
  m pocd/.style={okOrange, line width=0.7pt, mark=square*, mark size=1.1pt},
  m elite/.style={okGreen, line width=0.7pt, mark=diamond*, mark size=1.6pt},
  m ftd/.style={okPurple, line width=0.7pt, mark=triangle*, mark size=1.6pt,
    every mark/.append style={rotate=180}},
  m never/.style={okSky, line width=0.7pt, dash pattern=on 1.5pt off 1pt,
    mark=x, mark size=1.8pt, every mark/.append style={solid}},
}
\providecommand{\methodlegend}{%
  \addlegendimage{m never}\addlegendentry{Never forget}%
  \addlegendimage{m ftd}\addlegendentry{Fuse-then-decay}%
  \addlegendimage{m elite}\addlegendentry{ELite}%
  \addlegendimage{m pocd}\addlegendentry{POCD}%
  \addlegendimage{m pf}\addlegendentry{PF($\lambda$)}%
  \addlegendimage{m hiper}\addlegendentry{\textbf{HiPer (ours)}}}
\fi

\newcommand{\rThorOursNll}{0.265}
\newcommand{\rThorOursEce}{0.050}
\newcommand{\rThorOursBrier}{0.081}
\newcommand{\rThorOursAuroc}{0.911}

\newcommand{\rThorOursPrem}{0.102}

\newcommand{\rThorOursCalNll}{0.260}
\newcommand{\rThorOursCalEce}{0.050}

\newcommand{\rThorPfnNll}{0.274}

\newcommand{\rThorPfnCalNll}{0.263}

\newcommand{\rThorClsNll}{0.296}

\newcommand{\rThorClsAuroc}{0.887}

\newcommand{\rThorEnvNll}{0.273}

\newcommand{\rThorDinoNll}{0.270}

\newcommand{\rThorPfNll}{0.341}
\newcommand{\rThorPfEce}{0.045}

\newcommand{\rThorPfAuroc}{0.786}

\newcommand{\rThorPfPrem}{0.089}

\newcommand{\rThorPfCalEce}{0.041}

\newcommand{\rThorPocdNll}{0.402}

\newcommand{\rThorPocdStale}{1.000}
\newcommand{\rThorPocdCalNll}{0.326}

\newcommand{\rThorElitePrem}{0.118}

\newcommand{\rThorDraftStale}{0.838}

\newcommand{\rThorNoPriorNll}{0.326}

\newcommand{\rThorResetNll}{0.270}

\newcommand{\rThorHandGtNll}{0.390}

\newcommand{\rThorHandGtAuroc}{0.891}

\newcommand{\rThorHandAblNll}{0.545}

\newcommand{\rThorScenes}{120}

\newcommand{\rObcOursNll}{1.528}
\newcommand{\rObcOursEce}{0.649}

\newcommand{\rObcEnvNll}{1.277}

\newcommand{\rLoroOursNll}{0.498}

\newcommand{\rLoroOursCalNll}{0.487}

\newcommand{\rLoroPfNll}{0.427}

\newcommand{\rLoroPfCalNll}{0.446}

\newcommand{\rLoroNeverCalNll}{0.445}

\newcommand{\rFreshOursNll}{0.245}

\newcommand{\rOldOursNll}{0.270}

\newcommand{\rRscanOursNll}{0.310}
\newcommand{\rRscanOursEce}{0.122}
\newcommand{\rRscanOursBrier}{0.093}
\newcommand{\rRscanOursAuroc}{0.790}

\newcommand{\rRscanOursPrem}{0.467}
\newcommand{\rRscanOursStale}{0.792}
\newcommand{\rRscanOursCalNll}{0.311}

\newcommand{\rRscanPfnNll}{0.312}

\newcommand{\rRscanClsCalNll}{0.330}

\newcommand{\rRscanDinoNll}{0.328}

\newcommand{\rRscanPfEce}{0.098}

\newcommand{\rRscanPfAuroc}{0.522}

\newcommand{\rRscanPocdStale}{0.771}

\newcommand{\rRscanNoPriorNll}{0.369}

\newcommand{\rRscanResetNll}{0.313}

\newcommand{\rRscanScenes}{47}
\newcommand{\rRscanLabeled}{2107}

\newcommand{\dHeldScenes}{120}
\newcommand{\dHeldMoved}{3{,}560}

\newcommand{\dHeldDisappeared}{1{,}357}

\newcommand{\dObcScenes}{5}
\newcommand{\dObcSessions}{26}

\newcommand{\dRsTrainEnvs}{314}

\newcommand{\dRsValEnvs}{71}

\newcommand{\dRsTestEnvs}{47}

\newcommand{\dMapMs}{5.4}
\newcommand{\dRetrieveMs}{1.6}
\newcommand{\dObserveUs}{1.2}

\newcommand{\xNllFirstOurs}{0.291}
\newcommand{\xNllFirstPf}{0.381}
\newcommand{\xNllLastOurs}{0.283}
\newcommand{\xNllLastPf}{0.370}

\newcommand{\xGapMax}{0.090}

\newcommand{\xGapMin}{0.066}

\newcommand{\xCalGapMin}{0.040}
\newcommand{\xNllSessDraftMin}{0.835}
\newcommand{\xNllSessDraftMax}{0.912}
\newcommand{\xNllSessNeverMin}{1.967}
\newcommand{\xNllSessNeverMax}{2.904}
\newcommand{\xSeqN}{120}
\newcommand{\xWinNllPf}{116}

\newcommand{\xWinNllElite}{118}

\newcommand{\xWinNllPocd}{118}

\newcommand{\xLeaveLowOurs}{52}
\newcommand{\xStayHighOurs}{96.8}

\newcommand{\xConfErrOurs}{801}
\newcommand{\xLeaveLowPf}{38}
\newcommand{\xStayHighPf}{99.2}

\newcommand{\xConfErrPf}{1,558}

\newcommand{\xLeaveLowElite}{38}
\newcommand{\xStayHighElite}{98.5}

\newcommand{\xLeaveLowPocd}{1}
\newcommand{\xStayHighPocd}{100.0}

\newcommand{\xLeaveLowDraft}{76}
\newcommand{\xStayHighDraft}{61.5}

\newcommand{\xPredN}{42,374}

\newcommand{\xLoroPfSignP}{0.125}
\newcommand{\xLoroPfRoomsBetter}{4}
\newcommand{\xOracleNll}{0.288}
\newcommand{\xOracleAuroc}{0.913}
\newcommand{\xOracleBrier}{0.079}
\newcommand{\xOraclePfNll}{0.305}
\newcommand{\xValOursNll}{0.274}
\newcommand{\xValOursAuroc}{0.906}
\newcommand{\xValOursBrier}{0.083}

\newcommand{\xTsFewScenes}{4}
\newcommand{\xTsAllScenes}{60}
\newcommand{\xTsFewOurs}{0.300}
\newcommand{\xTsAllOurs}{0.274}
\newcommand{\xTsFewCls}{0.327}
\newcommand{\xTsAllCls}{0.307}
\newcommand{\xTsMidScenes}{20}
\newcommand{\xTsMidOurs}{0.283}
\newcommand{\xTsMidCls}{0.316}
\newcommand{\xRsKzeroPct}{44}
\newcommand{\xRsLongEnvs}{14}

\newcommand{\xRsLongDiff}{-0.001}
\newcommand{\xRsLongLo}{-0.020}
\newcommand{\xRsLongHi}{0.021}
\newcommand{\xRsDeparted}{23}
\newcommand{\xRsDepartedEnvs}{6}
\newcommand{\xRsAbsentObs}{46}
\newcommand{\xRsAbsentMiss}{7}
\newcommand{\xAsThorAll}{95}
\newcommand{\xAsThorReloc}{61}
\newcommand{\xAsThorNovel}{93}
\newcommand{\xAsThorGeoReloc}{3}
\newcommand{\xAsThorGeoNll}{0.394}
\newcommand{\xAsThorNoPoseNll}{0.267}
\newcommand{\xAsRsAll}{89}
\newcommand{\xAsRsReloc}{67}
\newcommand{\xAsRsNovel}{60}
\newcommand{\xAsRsGeoReloc}{13}
\newcommand{\xAsRsGeoNll}{0.334}

\newcommand{\xSensSelNll}{0.274}
\newcommand{\xSensKappaLow}{0.285}
\newcommand{\xSensNuHigh}{0.285}
\newcommand{\xSensFlatMax}{0.004}
\newcommand{\xSensStaleLow}{0.137}
\newcommand{\xSensPremLow}{0.045}
\newcommand{\xSensPremHigh}{0.955}
\newcommand{\xSensOtherMax}{0.004}
\newcommand{\xSensPfHigh}{0.291}

 \newcommand{\E}{\mathbb{E}}   \newcommand{\ie}{i.e.,\xspace} \newcommand{\hiper}{HiPer\xspace} \newcommand{\dtref}{\Delta_{\mathrm{ref}}}

\makeatletter \newcommand\captionof[1]{\def\@captype{#1}\caption} \let\hiper@maketitle\@maketitle \renewcommand{\@maketitle}{\hiper@maketitle}
\makeatother

\title{\LARGE \bf
Which Objects Stay? Learning Hierarchical Persistence Beliefs\\
for Lifelong Open-Vocabulary Object Maps}

\author{

Thanh Nguyen Canh\orcidlink{https://orcid.org/0000-0001-6332-1002},~\IEEEmembership{Graduate Student Member,~IEEE}, Haolan Zhang\orcidlink{https://orcid.org/0000-0001-6332-1002},~\IEEEmembership{Student Member,~IEEE}, and Nak Young Chong\orcidlink{https://orcid.org/0000-0001-5736-0769},~\IEEEmembership{Senior Member,~IEEE} 
\thanks{
This work was supported in part by JST SPRING, Japan Grant Number JPMJSP2102 and in part by the Asian Office of Aerospace Research and Development under Grant/Cooperative Agreement Award No. FA2386-25-1-4034. (Corresponding author: Nak Young Chong)}
\thanks{Thanh Nguyen Canh and Haolan Zhang are with the School of Information Science, Japan Advanced Institute of Science and Technology, Ishikawa 923-1211, Japan (email: thanhnc@jaist.ac.jp; haolan.z@jaist.ac.jp).}
\thanks{Nak Young Chong is with the School of Information Science, Japan Advanced Institute of Science and Technology, Ishikawa 923-1211, Japan, and also with the Department of Robotics, Hanyang University, Gyeonggi-do 15588, South Korea (email: nakyoung@jaist.ac.jp).}%
\thanks{Code and data are publicly available at \href{https://github.com/pairs-lab/HiPer}{HiPer}.}
\thanks{Digital Object Identifier 10.1109/LRA.2026.XXXXXXX.}
}

\begin{document}
\maketitle \thispagestyle{empty} \pagestyle{empty}

\begin{abstract}
Lifelong object maps must estimate whether an object remains at its previously observed location after it leaves view. Existing persistence filters rely on class-based survival priors, making them difficult to apply reliably in open-vocabulary settings. We introduce HiPer, a hierarchical persistence filter that derives persistence beliefs directly from data rather than semantic categories. For a new object, HiPer retrieves visually and geometrically similar training instances and estimates an initial survival rate from their observed histories, without requiring class labels. This estimate initializes a conjugate Beta belief over the object's survival probability, which is updated in closed form using detections and missed observations under Rosen’s persistence-filter framework. When an object is re-identified at a new location, its persistence belief is transferred with the object, enabling the map to learn object-specific mobility over time. Experiments on 120 held-out simulated sequences show that HiPer consistently achieves the best performance among competing methods. Retrieval-based priors outperform CLIP-derived label priors, and object-specific online learning provides further gains. Results on \rRscanScenes{} held-out 3RScan environments demonstrate that the proposed retrieval-based prior also outperforms state-of-the-art baselines in real-world settings.
\end{abstract}
\section{Introduction}
\label{sec:intro}
Robots that repeatedly navigate the same rooms, for example to retrieve or organize objects, rely on long-term object maps built from past observations~\cite{lee2026ltmem,saavedra2026predictive}. These maps are useful for object search, scene rearrangement, and relocalization~\cite{bogenberger2026glasses,weihs2021roomr,wald2019rio}, but only when they remain consistent with the environment’s current state. After each mapping session, the robot must estimate whether an object remains at its previously observed location. Some objects, such as mugs, chairs, and boxes, are frequently moved, whereas others, such as counters and refrigerators, are largely stationary. A map that retains all objects indefinitely accumulates stale information, degrading object search and relocalization performance. Conversely, a map that removes objects too aggressively may discard objects that are merely outside the robot’s field of view. Lifelong mapping systems therefore monitor changes in the environment and remove objects that are no longer present~\cite{gil2025elite,oh2026oasis,xu2026focus}. For individual objects, this requires reasoning under uncertainty by balancing prior expectations about how long similar objects typically remain in place with evidence gathered from observations of the object itself.

\begin{figure}
    \centering
    \includegraphics[width=\linewidth]{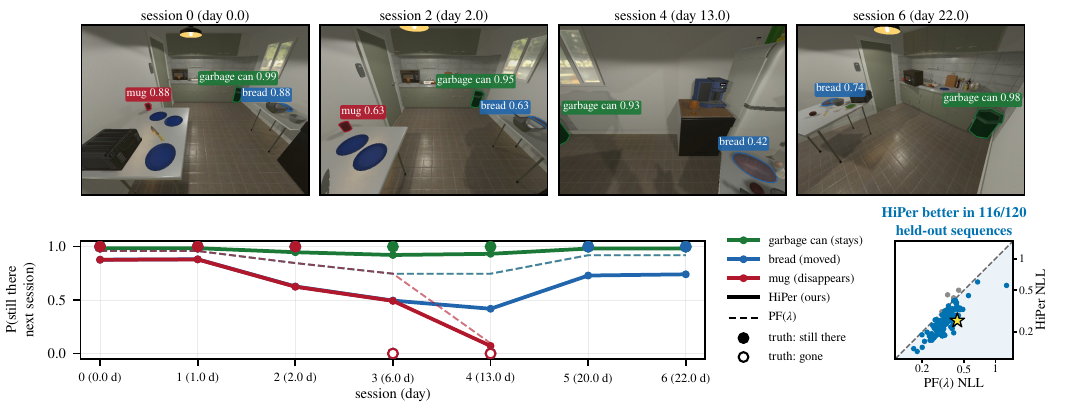}
      \caption{\textbf{Which objects stay?} Top: four sessions of \hiper's best held-out THOR-Lifelong sequence. Bottom: predicted probability that an object stays in place until the next session for \hiper (solid) and PF($\lambda$)~\cite{rosen2016persistence} (dashed); dots: ground truth (open: gone). Right: negative log-likelihood (NLL) of all \xSeqN{} held-out sequences; \hiper attains the lower NLL in \xWinNllPf{} of them.}
    \label{fig:teaser}
\end{figure}

Probabilistic persistence models provide a principled framework for reasoning about object persistence under uncertainty. Rosen \textit{et al}.~\cite{rosen2016persistence} compute the exact posterior probability that a feature still exists using a survival prior and a noisy detection model. More recent object-level approaches maintain Bayesian beliefs over object persistence~\cite{qian2022pocd}, and open-vocabulary systems often derive these beliefs from language-model estimates associated with an object's semantic category~\cite{bogenberger2026glasses,adkins2026presist,lee2026ltmem,saavedra2026predictive}. Despite these advances, three gaps remain. First, the prior is often weak. Persistence is commonly predicted from class-level statistics, yet object categories provide only a coarse indication of how long a particular object is likely to remain in place. This problem is compounded by the noise of open-vocabulary recognition: in our dataset, CLIP's zero-shot label matches the exact object type for only about one-third of tracks. Second, a prior is only an initial estimate. Objects belonging to the same category can behave very differently. For instance, while mugs are generally movable, a particular mug may remain in the same location for months. A fixed-rate exponential model cannot learn such object-specific persistence because it is memoryless. Rosen \textit{et al}. mitigate this limitation by marginalizing a log-uniform hyperprior over the survival rate~\cite{rosen2016persistence}, allowing persistence beliefs to strengthen as an object continues to survive. However, the resulting prior remains generic and is shared by all objects, irrespective of their appearance or geometry. Third, persistence is commonly modeled at fixed locations rather than for object identities, making it difficult to capture the mobility patterns of objects that are repeatedly moved and later re-observed elsewhere.

We close these gaps with \hiper, a \emph{hierarchical persistence filter} for open-vocabulary object maps (Figs.~\ref{fig:teaser} and~\ref{fig:overview}). \hiper represents persistence at three levels. First, it maintains a memory of training objects together with their observed histories of persistence and relocation. Second, for a newly observed object, \hiper retrieves the most similar training objects in terms of appearance and size and uses their histories to estimate a survival prior, without relying on semantic class labels. Third, each object maintains its own persistence belief, which is updated from detections and missed observations over time. When an object is re-identified at a new location, its persistence belief is transferred with the object, allowing the system to learn object-specific mobility patterns. We evaluate \hiper on held-out simulated sequences with ground-truth persistence labels and on real-world 3RScan environments~\cite{wald2019rio}, comparing against Rosen’s persistence filter, ELite~\cite{gil2025elite}, POCD~\cite{bogenberger2026glasses}, and two map-maintenance baselines. The main contributions of this paper are as follows:
\begin{enumerate}
\item \emph{Class-free retrieval of persistence priors.} 
We estimate the survival rate of a new object from the recorded histories of training objects that are most similar in appearance and geometry (\textit{e.g.}, size, height, and mask area). The estimate is computed using a weighted, interval-censored exponential model with shrinkage toward a global rate, eliminating the need for class labels, language models, or manually specified lifetime tables.
\item \emph{
Object-specific Bayesian persistence learning.}
The retrieved survival rate initializes a Beta prior over the object's persistence probability. This belief is updated in closed form using detections and missed observations under Rosen's detection model. The proposed formulation reduces to Rosen's persistence filter in the limit of an infinitely strong prior, while allowing individual objects to become more or less persistent based on their observed behavior.
\item \emph{
Persistence tied to object identity rather than location.} 
When an object is re-identified at a new location, the learned persistence belief is transferred with the object. The previous location is treated as a disappearance event, allowing the map to learn the mobility of individual objects over time instead of restarting persistence estimation at each new position.
\end{enumerate}

\section{Related Work}
\label{sec:related}

\noindent\textbf{Persistence filtering.} Rosen \textit{et al.}~\cite{rosen2016persistence} introduced the persistence filter, a recursive Bayesian belief over the survival time of a feature under a survival prior and a detector with known miss and false-alarm rates. They suggested class-conditional hazard rates and a hyperprior over the rate as extensions. Perpetua~\cite{saavedra2025perpetua} combines persistence and emergence filters whose parameters are learned offline by expectation maximization (EM), and predictive scene graphs~\cite{saavedra2026predictive} add language-model priors on object--receptacle relations. ELite~\cite{gil2025elite} maintains within-session and cross-session ephemerality for lifelong LiDAR mapping with scalar Bayesian fusion. \hiper retains Rosen's detection model and learns a survival rate per object under a prior retrieved by appearance, online and in closed form.

\noindent\textbf{Object-level change and semi-static maps.} Probabilistic object-level change detection (POCD)~\cite{qian2022pocd} maintains a joint Gaussian geometric-change and Beta stationarity belief per object with closed-set class priors. Bogenberger \textit{et al.}~\cite{bogenberger2026glasses} extend it to open-vocabulary objects with a language-model prior on whether an object is static or dynamic and an out-of-view decay. FOCUS~\cite{xu2026focus} fuses geometric and appearance evidence in a per-object change belief, and LT-Mem~\cite{lee2026ltmem} carries a language-initialized object volatility across sessions. These methods threshold a mean stationarity or volatility under a fixed prior. In contrast, we learn a survival rate per object under a prior estimated from data, and we report calibration, which these works do not.

\noindent\textbf{Open-vocabulary priors on persistence.} DualMap~\cite{jiang2025dualmap} separates anchor and volatile objects by their CLIP similarity to class lists, without a probabilistic model. PreSIST~\cite{adkins2026presist} predicts instance-level survival-time priors with a vision-language model and feeds them to a persistence filter. Its baseline with CLIP and a class lookup corresponds to our CLIP class-rate variant. \hiper instead estimates an instance-level prior from the recorded histories of training objects that look alike, without a language model. VLMM~\cite{ghosh2026vlmm} attaches calibrated movability attributes to 3D maps, and LatentBKI~\cite{wilson2025latentbki} models the uncertainty of open-vocabulary embeddings in voxel maps.

\noindent\textbf{Cross-session object association.} BYE~\cite{huang2025bye} trains a per-scene point-cloud encoder to re-identify relocated objects, CLOVER~\cite{adkins2025clover} learns viewpoint-invariant object descriptors, and OASIS-Map~\cite{oh2026oasis} matches objects across revisits by dense semantic correspondence. We associate objects with foundation-model embeddings~\cite{radford2021clip,oquab2024dinov2} and carry each object's persistence belief across its relocations.

\section{HiPer: Hierarchical Persistence Beliefs}
\label{sec:method}

\begin{figure*}[t]
\centering
\includegraphics[width=\textwidth]{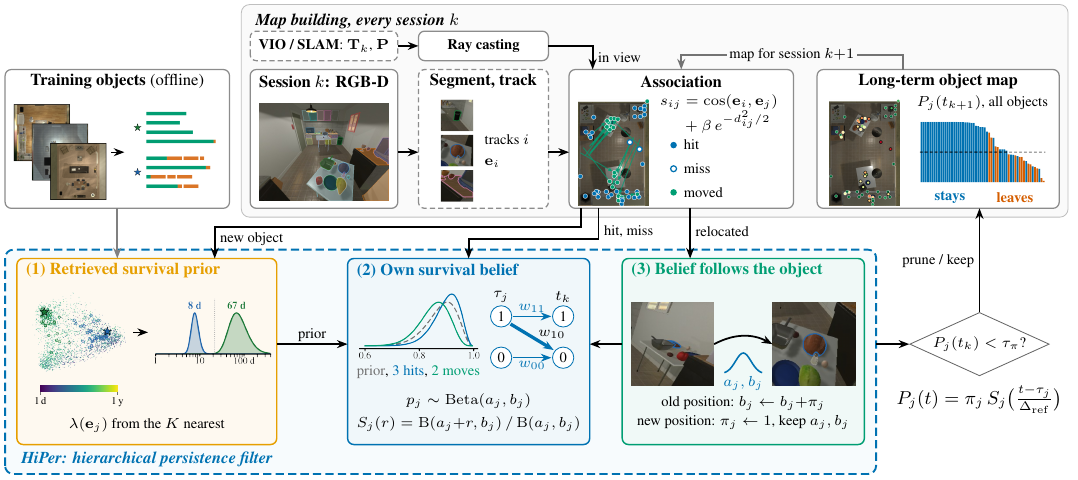}
\caption{Overview of \hiper on a held-out sequence. Association turns each session into one piece of evidence per map object (hit, miss or moved); ray casting decides which unmatched map objects are in view and thus missed. (1)~The survival prior of a new object is retrieved from the $K$ nearest training objects (2-D projection of their DINOv2 embeddings, colored by observed lifetime; stars: the garbage can and the bread). (2)~Each object keeps a Beta belief over its own survival, updated in closed form (the bread: prior, after three hits and after two relocations). (3)~At a relocation, the belief moves with the object. Objects whose predicted presence falls below $\tau_\pi$ are pruned. Top right: predicted presence of all map objects at the next session.}
\label{fig:overview}
\end{figure*}

\subsection{Problem Formulation}
\label{sec:problem}

A robot revisits an environment over multiple sessions $k=0,1,\dots$ at times $t_0<t_1<\dots$, with intervals ranging from hours to weeks. During each session, it observes object instances that are segmented and tracked within the session, together with RGB-D images, camera pose estimates and associated uncertainties, and appearance embeddings. The robot maintains a long-term object map in which each object $j$ is represented by a position $\mathbf{x}_j$, covariance $\bm\Sigma_j$, appearance embedding $\mathbf{e}_j$, open-vocabulary class label $c_j$, and a persistence belief. After session $k$, the map estimates, for every object, the probability that the object will remain at its mapped location until the next visit at time $t_{k+1}$. Based on these estimates, the system must also decide which objects to retain and which to remove from the map. We evaluate persistence prediction using proper scoring rules and calibration metrics, and map maintenance using two complementary measures: premature pruning, where objects are removed despite still being present, and staleness, where objects are retained after they have disappeared. An overview of the proposed method is shown in Fig.~\ref{fig:overview}.

\subsection{Survival Prior Retrieved by Appearance}
\label{sec:prior}

The survival prior represents how long an object is expected to remain at its previously observed location. We parameterize this prior by a survival rate $\lambda$, defined as the inverse of the expected stay duration. Rather than deriving $\lambda$ from a semantic class label, we estimate it from training objects that are similar in appearance and geometry. For each training object $i$, we store its mean DINOv2~\cite{oquab2024dinov2} and CLIP embeddings, a standardized geometry descriptor, and its persistence history. The geometry descriptor includes the logarithms of the sorted object extents, the centroid height, and the logarithm of the mean mask area. Each consecutive pair of sessions contributes either an exposure interval $d$, if the object remains present, or an interval-censored disappearance event within $d$, if the object vanishes, moves, or reappears elsewhere. Let $T_i$ denote the total exposure time and $d_{i\ell}$ the intervals ending in disappearance events. For a newly created map object with features $\mathbf{e}$, similarity to training object $i$ is measured by combining DINOv2 similarity, CLIP similarity, and geometric similarity:

\begin{equation}
  \delta_i = \frac{\alpha\,(1-s^{D}_i) + (1-\alpha)\,r_c\,(1-s^{C}_i)
           + w_g\, r_g\, \delta^{G}_i}{1+w_g},
  \label{eq:distance}
\end{equation}
where $s^{D}_i$ and $s^{C}_i$ are cosine similarities in the DINOv2 and CLIP feature spaces, and $\delta^{G}_i$ measures geometric dissimilarity. Missing features are ignored. Scale factors $r_c$ and $r_g$ normalize the contributions of the CLIP and geometry terms, while $\alpha=0.75$ and $w_g=1$ are selected on a validation set. The $K$ nearest training objects are assigned weights $\omega_i = e^{-(\delta_i-\delta_{\min})/\tau}$, where $\delta_{\min}$ is the smallest retrieval distance among the selected neighbors. Their histories are used to estimate a retrieval-based survival rate:
\begin{equation}
  \begin{aligned}
  \lambda(\mathbf{e}) = \arg\max_{\lambda}\;
  &\mathrm{Ga}\bigl(\lambda;\,1{+}\nu,\,\nu/\lambda_0\bigr)\\
  &\times\prod_{i}\Bigl[e^{-\lambda T_i}\prod_{\ell}
  \bigl(1-e^{-\lambda d_{i\ell}}\bigr)\Bigr]^{\omega_i}.
  \end{aligned}
  \label{eq:retrieval}
\end{equation}

\begin{figure}[t]
  \centering
\begin{tikzpicture}
\begin{groupplot}[
    group style={group size=2 by 1, horizontal sep=0.3cm,
                 y descriptions at=edge left},
    scale only axis, width=0.4\columnwidth, height=2.4cm,
    xmin=0.5, xmax=14.5, ymin=0.3, ymax=1.04,
    xtick={1,4,7,10,13}, ytick={0.4,0.6,0.8,1},
    xlabel={session (day)}, label style={font=\scriptsize},
    tick label style={font=\tiny},
    title style={font=\scriptsize, yshift=-1.4mm},
    grid=major, grid style={black!12}, axis line style={black!60},
    legend style={font=\tiny, draw=black!25, legend columns=3,
                  /tikz/every even column/.append style={column sep=5pt}},
    legend cell align=left]
  \nextgroupplot[title={mug that stays},
                 ylabel={$P_j(t_k{+}1\,\mathrm{day})$},
                 legend to name=belieflegend]
    \methodlegend
    \addplot[m never, forget plot] table[x=day, y=never_a, col sep=comma] {figures/data/hpf_belief.csv};
    \addplot[m ftd, forget plot] table[x=day, y=ftd_a, col sep=comma] {figures/data/hpf_belief.csv};
    \addplot[m elite, forget plot] table[x=day, y=elite_a, col sep=comma] {figures/data/hpf_belief.csv};
    \addplot[m pocd, forget plot] table[x=day, y=pocd_a, col sep=comma] {figures/data/hpf_belief.csv};
    \addplot[m pf, forget plot] table[x=day, y=pf_pred, col sep=comma] {figures/data/hpf_belief.csv};
    \addplot[m hiper, forget plot] table[x=day, y=a_pred, col sep=comma] {figures/data/hpf_belief.csv};
  \nextgroupplot[title={mug moved every second day}]
    \addplot[m never, forget plot] table[x=day, y=never_b, col sep=comma] {figures/data/hpf_belief.csv};
    \addplot[m ftd, forget plot] table[x=day, y=ftd_b, col sep=comma] {figures/data/hpf_belief.csv};
    \addplot[m elite, forget plot] table[x=day, y=elite_b, col sep=comma] {figures/data/hpf_belief.csv};
    \addplot[m pocd, forget plot] table[x=day, y=pocd_b, col sep=comma] {figures/data/hpf_belief.csv};
    \addplot[m pf, forget plot] table[x=day, y=pf_b, col sep=comma] {figures/data/hpf_belief.csv};
    \addplot[m hiper, forget plot] table[x=day, y=b_pred, col sep=comma] {figures/data/hpf_belief.csv};
\end{groupplot}
\path (group c1r1.north east) -- node[above=13pt]
  {\pgfplotslegendfromname{belieflegend}} (group c2r1.north west);
\end{tikzpicture}
  \caption{Probability of remaining in place for one more day, for two mugs with the same prior (lifetime of 2 days) observed daily ($\kappa_0=32$, $P_M=0.2$, $P_F=0.001$); left: the mug stays, right: it moves every second day. \hiper's prediction rises for the mug that stays and falls for the one that moves. PF($\lambda$) and fuse-then-decay predict the same for both, and ELite and POCD approach certainty for the mug that stays.}
  \label{fig:belief}
\end{figure}
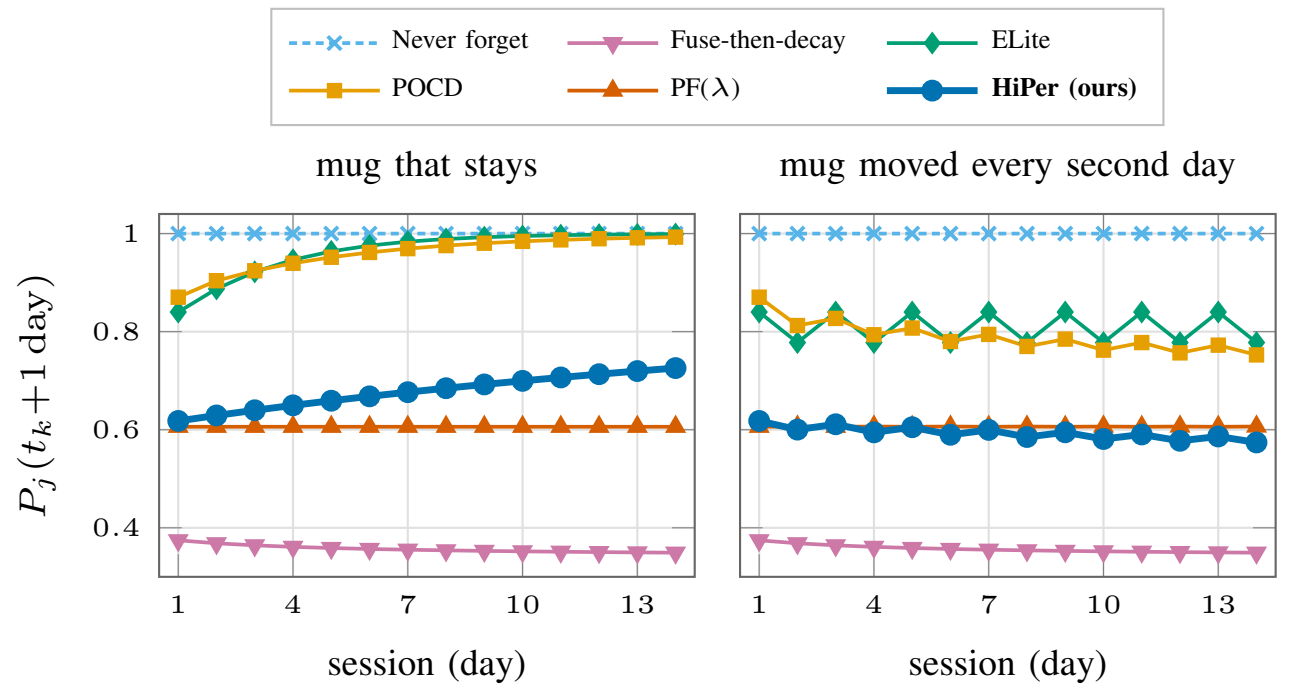

This corresponds to a weighted maximum-likelihood estimate under an exponential survival model, regularized toward the global rate $\lambda_0$ through a Gamma prior equivalent to $\nu$ pseudo-observations. The optimum is found efficiently by bisection. As a baseline, the same estimator can be applied at the class level by assigning nonzero weights only to training objects with the same semantic label. We refer to this alternative as the CLIP class-rate prior. We also compare against a manually specified lifetime table. Let $p_j$ denote the probability that object $j$ remains at its location over a one-day reference interval $\dtref$. The retrieved survival rate defines the prior belief

\begin{equation}
  p_j \sim \mathrm{Beta}\bigl(\kappa_0 m_j,\ \kappa_0 (1-m_j)\bigr),
  \qquad m_j = e^{-\lambda(\mathbf{e}_j) \dtref},
  \label{eq:prior}
\end{equation}
where $\kappa_0$ controls the prior strength. This yields a three-level hierarchy consisting of training objects, the retrieved neighborhood, and the object-specific belief. Unlike a fixed survival rate, a finite $\kappa_0$ enables adaptation over time: objects that repeatedly remain in place become increasingly persistent, whereas objects that frequently move become less so (Fig. \ref{fig:belief}).

\begin{table}[t]
\caption{Hand-set class-conditioned lifetimes $1/\lambda_c$}
\label{tab:priors}
\centering\small
\begin{tabular*}{\columnwidth}{@{\extracolsep{\fill}}l||c@{}}
\toprule
Types (examples) & $1/\lambda_c$ \\
\midrule
Perishables, newspaper (apple, bread, egg)      & 1 day \\
Handled daily (phone, keys, mug, cup, plate)    & 1--3 days \\
Weekly items (soap, pillow, book, chair)        & 5--14 days \\
Decor and light furniture (vase, lamp, plant)   & 30--180 days \\
Large furniture, appliances (sofa, bed, TV)     & 1--5 years \\
Fixtures (counter, cabinet, sink, fridge)          & 10 years \\
\bottomrule
\end{tabular*}
\end{table}

\subsection{Hierarchical Persistence Filter}
\label{sec:filter}

Each object $j$ maintains a presence probability $\pi_j$ at the time $\tau_j$ of its most recent informative update, together with Beta parameters $(a_j,b_j)$ representing its persistence belief. Over a time interval $\Delta t$, the probability that the object remains at its current location is modeled as $p_j^{\,r}$, where $r=\Delta t/\dtref$. Assuming exponential survival, $p_j$ corresponds to a survival rate of $-\ln p_j/\dtref$. Under the Beta belief, the expected survival probability over $r$ reference intervals is

\begin{equation}
  S_j(r) = \E\bigl[p_j^{\,r}\bigr]
         = \frac{\mathrm{B}(a_j+r,\,b_j)}{\mathrm{B}(a_j,\,b_j)} .
  \label{eq:survival}
\end{equation}

During each session $k$, an object may generate one of three outcomes: a hit, a miss, or no evidence. A hit occurs when the object is confidently re-observed at its mapped location. A miss occurs when the location is visible but no associated observation is found. No evidence is recorded when the location lies outside the field of view. Following Rosen's persistence model~\cite{rosen2016persistence}, detections are characterized by a miss probability $P_M$ and a false-alarm probability $P_F$. Consequently, the likelihoods of a hit are $(L_1,L_0)=(1-P_M,\,P_F)$ under the hypotheses that the object is present or absent, respectively, while a miss has likelihoods $(P_M,\,1-P_F)$. Given an elapsed interval $r=(t_k-\tau_j)/\dtref$, the filter first predicts the object's survival and then incorporates the new observation. This yields the two-slice posterior

\begin{equation}
  w_{11}\propto \pi_j S_j L_1,\quad
  w_{10}\propto \pi_j (1-S_j) L_0,\quad
  w_{00}\propto (1-\pi_j) L_0 ,
  \label{eq:twoslice}
\end{equation}
corresponding to the probabilities that the object was present at $\tau_j$ and remains present, was present but disappeared, or was already absent, respectively. After normalization, the object belief is updated as
\begin{equation}
  \begin{aligned}
  \pi_j &\leftarrow w_{11}, &\quad a_j &\leftarrow a_j + r\,w_{11},\\
  \tau_j &\leftarrow t_k,  &\quad b_j &\leftarrow b_j + w_{10}.
  \end{aligned}
  \label{eq:update}
\end{equation}
Intuitively, evidence that an object continues to persist increases $a_j$, strengthening the belief that it tends to remain in place, whereas evidence of disappearance increases $b_j$, indicating greater mobility. As a result, each object gradually learns its own persistence behavior from experience.

Each object $j$ maintains a presence probability $\pi_j$ at the time $\tau_j$ of its most recent informative update and Beta parameters $(a_j,b_j)$ representing its persistence belief. For a certain observation ($w_{11}\in\{0,1\}$), the update of $a_j$ is the exact conjugate update because the survival likelihood $p_j^{\,r}$ is conjugate to the Beta prior. Under uncertain evidence, the update adds the expected sufficient statistic, yielding an assumed-density approximation. The update of $b_j$ follows from the first-order approximation of the disappearance likelihood $1-p_j^{\,r}$. Sessions that provide no evidence leave the belief unchanged. Consequently, exposure accumulates implicitly and is accounted for only when the object is observed again. When data association re-identifies an object at a new location (Sec. \ref{sec:assoc}), the previous location is treated as a disappearance event, and $b_j \leftarrow b_j+\pi_j$. The object is then reinitialized at its new location with $\pi_j\leftarrow 1$, while the persistence belief $(a_j,b_j)$ is preserved. In contrast to location-based persistence filters, which restart from the prior after relocation, \hiper retains the object's accumulated persistence history and thereby learns object-specific mobility patterns. The probability that object $j$ remains at its mapped location at a future time $t$ is

\begin{equation}
  P_j(t) = \pi_j\, S_j\bigl((t-\tau_j)/\dtref\bigr),
  \label{eq:predict}
\end{equation}
where $\pi_j$ captures the current belief that the object is present and $S_j(\cdot)$ models its expected survival over the elapsed time interval. Here, $\tau_j$ denotes the time of the last informative update of object $j$, while $t$ denotes an arbitrary future time at which the persistence probability is evaluated.

\subsection{Environment Rate and Map Maintenance}
\label{sec:env}

A persistence prior learned in one environment may not transfer well to another environment with different dynamics. To account for this, \hiper optionally introduces a top-level adaptation layer that learns an environment-specific hazard multiplier $\gamma$. This multiplier rescales the survival model of every object, replacing $S_j(r)$ with $S_j(\gamma r)$, and is estimated online from the collective behavior of all objects in the map. For each update, the filter accumulates the expected number of disappearance events, $H \mathrel{+}= -\pi_j\ln S_j(r)$, and the corresponding observed count, $O \mathrel{+}= w_{10}$, with relocations contributing $\pi_j$ additional disappearance events. Assuming a Gamma prior centered at one, the hazard multiplier is given by the mean of the posterior distribution, $\gamma \sim \mathrm{Ga}(\alpha_\gamma+O,\,\alpha_\gamma+H)$. This adaptation allows the persistence model to automatically adjust to environments whose objects move more or less frequently than those observed during training. Finally, after each session, an object is removed from the map if its predicted presence probability falls below a threshold, $P_j(t_k)<\tau_\pi$, where $\tau_\pi$ controls the trade-off between prematurely removing objects and retaining stale map entries.

\subsection{Cross-Session Association}
\label{sec:assoc}

Within each session, object tracks $i$ are associated with map objects $j$ using Hungarian assignment on the similarity score
\begin{equation}
  s_{ij} = \cos(\mathbf{e}_i,\mathbf{e}_j) + \beta\, e^{-d^2_{ij}/2},
  \label{eq:assoc}
\end{equation}
where $\mathbf{e}$ denotes the mean DINOv2~\cite{oquab2024dinov2} embedding of an object's image crops and $d^2_{ij}$ is the Mahalanobis distance computed using the covariance $\bm\Sigma_j+\mathbf{P}+\mathbf{R}_i$. Here, $\mathbf{\Sigma}_j$ is the object's accumulated localization uncertainty, $\mathbf{P}$ is the session pose covariance, and $\mathbf{R}_i$ is the measurement covariance of the track. A track-object pair is associated when $s_{ij}\ge\tau_m$. Associated pairs with $d^2_{ij}>\chi^2_3(0.99)$ are treated as relocations, while those within this gate are considered hits and used to update the object's position and covariance through fusion. Unmatched tracks initialize new map objects. Unmatched map objects whose locations are visible in the current session, as determined by ray-casting into the depth images and visibility checks, are assigned misses. All parameters are selected on a validation set. Association parameters are optimized for matching accuracy and include the appearance-spatial weighting factor $\beta=0.25$ and matching threshold $\tau_m=0.5$. Persistence prediction uses $\kappa_0=32$, $P_M=0.2$, $P_F=0.001$, $\tau_\pi=0.5$, $K=30$, $\tau=0.05$, and $\nu=0.25$.

\section{Experiments}
\label{sec:experiments}

\subsection{Experimental Setup}
\label{sec:setup}

\noindent\textbf{Datasets.} \emph{THOR-Lifelong} contains eight-session sequences in the AI2-THOR~\cite{kolve2017ai2thor} floor plans of RoomR~\cite{weihs2021roomr}, with sessions 0.5 to 7 days apart and exact ground truth. Objects move, disappear and reappear at rates set by their affordance, not their class. Priors are learned on 60 floor plans and methods are tuned on 20; the test data are three held-out sets of 40 sequences on the 40 test floor plans (\dHeldScenes{} sequences, \dHeldMoved{} moves, \dHeldDisappeared{} removals). \emph{3RScan}~\cite{wald2019rio} provides annotated rescans of real indoor environments, treated as sessions one day apart (moved: displaced by at least 0.2\,m). Its official test labels are hidden, so its validation split (\dRsTestEnvs{} environments) is our test set, and its training split provides \dRsTrainEnvs{} environments for the priors and \dRsValEnvs{} for tuning. \emph{ObChange}~\cite{langer2022obchange} contains \dObcSessions{} robot visits to \dObcScenes{} rooms. As in BYE~\cite{huang2025bye}, segmentation and tracking within a session are oracle. Camera poses are the ground truth with simulated drift on THOR and the dataset alignment on 3RScan and ObChange. Embeddings come from CLIP ViT-B/32 and DINOv2 ViT-B/14.

\noindent\textbf{Baselines and protocol.} All methods replay the same evidence. The baselines are PF($\lambda$), Rosen's persistence filter with a single tuned rate~\cite{rosen2016persistence}; ELite~\cite{gil2025elite}; the POCD belief, ported from the code of~\cite{bogenberger2026glasses}; fuse-then-decay, \ie ELite's fusion followed by a persistence-filter decay~\cite{deng2023ltvs}; and a map that never forgets, the static-world assumption~\cite{cadena2016past}. Every method is tuned on validation (prediction by NLL, pruning by the premature-prune and stale rates) and is also reported with a calibration map fitted on validation. The three THOR held-out sets were each evaluated once after freezing and are pooled (calibration and pooling were adopted after the first set); 3RScan was evaluated once, with a protocol fixed before download. ObChange and two earlier THOR test splits were analyzed during development; on the latter, \hiper also has the lowest NLL (\rOldOursNll{} and \rFreshOursNll). The environment rate, the retrieval key and cross-fitting were added after inspecting ObChange. We report NLL, Brier score, expected calibration error (ECE) and AUROC (AUC in the tables) as mean $\pm$ standard deviation over sequences (all methods are deterministic). A difference is significant if the 95\% interval of a paired bootstrap over sequences excludes zero. Bold and underlined values mark the best and second-best means.

\begin{table*}[t]
\caption{Results on the 120 held-out THOR-Lifelong sequences (mean $\pm$ standard deviation over sequences). Calibrated: with a calibration map fitted on validation. Bold: best; underlined: second best.}
\label{tab:persistence}
\centering\scriptsize\setlength{\tabcolsep}{3.0pt}
\begin{tabularx}{\linewidth}{@{}l||YYYY|YY|YY@{}}
\toprule
\multirow{2}{*}{Method}
 & \multicolumn{4}{c|}{Prediction} & \multicolumn{2}{c|}{Calibrated} & \multicolumn{2}{c}{Map maintenance}\\
 & NLL$\downarrow$ & Brier$\downarrow$ & ECE$\downarrow$ & AUC$\uparrow$ & NLL$\downarrow$ & ECE$\downarrow$ & Prem.$\downarrow$ & Stale$\downarrow$\\
\midrule
Never forget~\cite{cadena2016past} & 2.470\,{\tiny$\pm$0.772} & 0.179\,{\tiny$\pm$0.056} & 0.179\,{\tiny$\pm$0.056} & 0.500\,{\tiny$\pm$0.000} & 0.344\,{\tiny$\pm$0.076} & \textbf{0.038}\,{\tiny$\pm$0.025} & -- & 1.000\,{\tiny$\pm$0.000}\\

Fuse-then-decay~\cite{deng2023ltvs} & 0.879\,{\tiny$\pm$0.299} & 0.255\,{\tiny$\pm$0.061} & 0.328\,{\tiny$\pm$0.075} & 0.769\,{\tiny$\pm$0.081} & 0.325\,{\tiny$\pm$0.079} & 0.058\,{\tiny$\pm$0.024} & 0.146\,{\tiny$\pm$0.206} & 0.838\,{\tiny$\pm$0.182}\\

ELite scalar~\cite{gil2025elite} & 0.365\,{\tiny$\pm$0.099} & 0.104\,{\tiny$\pm$0.029} & 0.064\,{\tiny$\pm$0.026} & 0.773\,{\tiny$\pm$0.073} & 0.341\,{\tiny$\pm$0.079} & 0.051\,{\tiny$\pm$0.023} & 0.118\,{\tiny$\pm$0.146} & \underline{0.011}\,{\tiny$\pm$0.041}\\

POCD~\cite{bogenberger2026glasses} & 0.402\,{\tiny$\pm$0.092} & 0.127\,{\tiny$\pm$0.034} & 0.076\,{\tiny$\pm$0.036} & 0.784\,{\tiny$\pm$0.073} & 0.326\,{\tiny$\pm$0.077} & 0.059\,{\tiny$\pm$0.021} & -- & 1.000\,{\tiny$\pm$0.000}\\

PF($\lambda$)~\cite{rosen2016persistence} & \underline{0.341}\,{\tiny$\pm$0.113} & \underline{0.097}\,{\tiny$\pm$0.028} & \textbf{0.045}\,{\tiny$\pm$0.025} & \underline{0.786}\,{\tiny$\pm$0.062} & \underline{0.318}\,{\tiny$\pm$0.076} & \underline{0.041}\,{\tiny$\pm$0.022} & \textbf{0.089}\,{\tiny$\pm$0.133} & \textbf{0.008}\,{\tiny$\pm$0.033}\\

HiPer (ours) & \textbf{0.265}\,{\tiny$\pm$0.078} & \textbf{0.081}\,{\tiny$\pm$0.025} & \underline{0.050}\,{\tiny$\pm$0.020} & \textbf{0.911}\,{\tiny$\pm$0.040} & \textbf{0.260}\,{\tiny$\pm$0.076} & 0.050\,{\tiny$\pm$0.020} & \underline{0.102}\,{\tiny$\pm$0.128} & 0.013\,{\tiny$\pm$0.069}\\

\bottomrule
\end{tabularx}
\end{table*}

\begin{figure}[t]
  \centering
  \includegraphics[width=\columnwidth]{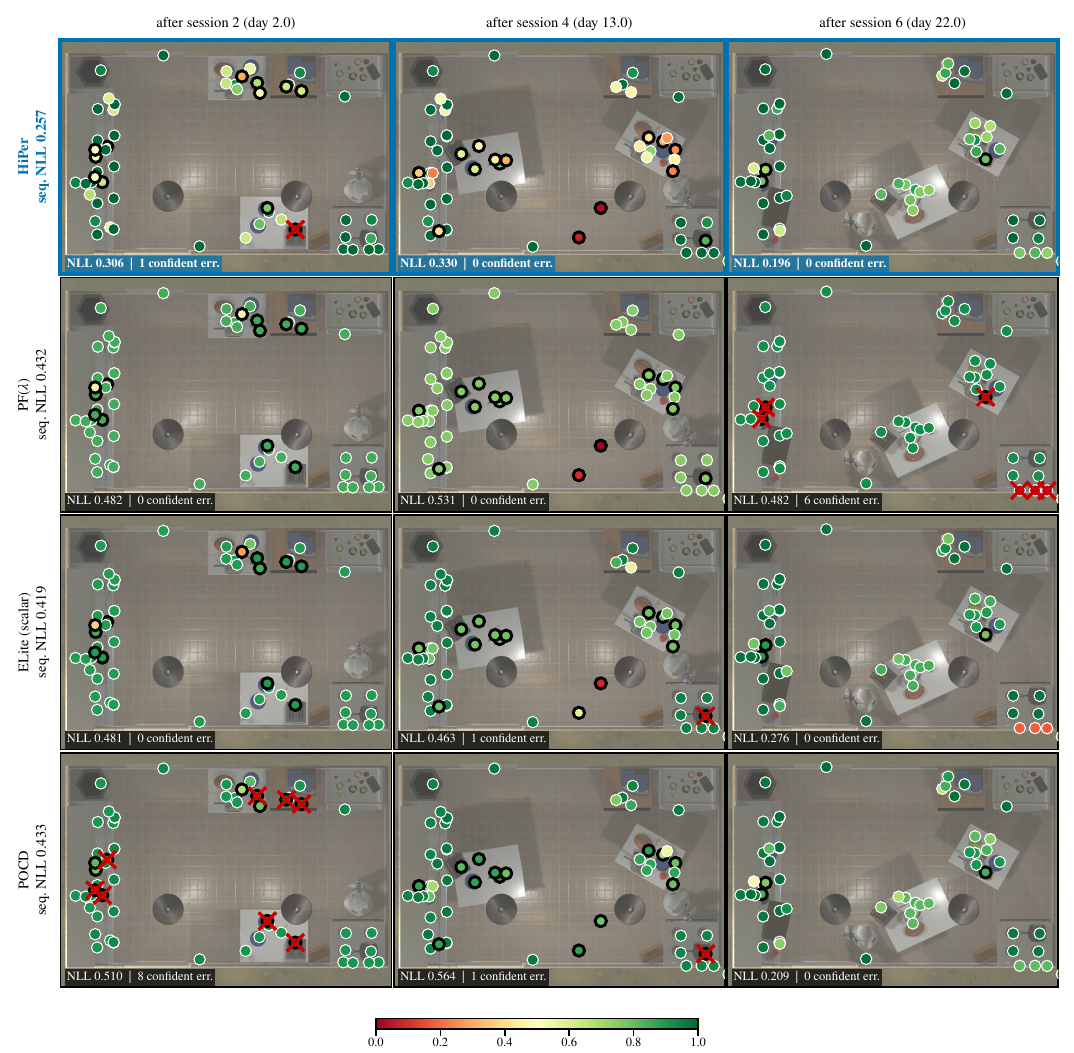}
  \caption{Qualitative results on THOR-Lifelong sequence. Color: predicted probability of remaining in place until the next session; black ring: the object leaves; red cross: confident error. Each panel reports the NLL of the session (bold: best) and the number of confident errors; the row labels give the NLL of the sequence.}
  \label{fig:qualitative}
\end{figure}

\subsection{Persistence Prediction}

\begin{figure}[t]
  \centering
\begin{tikzpicture}
\begin{axis}[width=\columnwidth, height=4.7cm, xmin=0, xmax=1.03, ymin=0, ymax=1,
    xtick={0,0.2,...,1},
    xlabel={predicted $P$(still there at next session)},
    ylabel={observed frequency}, label style={font=\scriptsize},
    tick label style={font=\tiny}, grid=major, grid style={black!12},
    legend style={font=\tiny, at={(0.5,1.03)}, anchor=south,
                  draw=black!25, legend columns=3,
                  /tikz/every even column/.append style={column sep=5pt}},
    legend cell align=left]
  \methodlegend
  \addplot[black!40, dashed, forget plot] coordinates {(0,0) (1,1)};
  \addplot[m ftd, forget plot] table[x=mean_pred, y=frac_pos, col sep=comma] {figures/data/reliability_draft_v1.csv};
  \addplot[m elite, forget plot] table[x=mean_pred, y=frac_pos, col sep=comma] {figures/data/reliability_elite.csv};
  \addplot[m pocd, forget plot] table[x=mean_pred, y=frac_pos, col sep=comma] {figures/data/reliability_pocd.csv};
  \addplot[m pf, forget plot] table[x=mean_pred, y=frac_pos, col sep=comma] {figures/data/reliability_pf.csv};
  \addplot[m never, only marks, mark size=2.4pt, line width=1pt, forget plot]
    table[x=mean_pred, y=frac_pos, col sep=comma] {figures/data/reliability_always.csv};
  \addplot[m hiper, forget plot]
    table[x=mean_pred, y=frac_pos, col sep=comma] {figures/data/reliability_ours.csv};
\end{axis}
\end{tikzpicture}
  \caption{Reliability diagram on the held-out sequences (15 bins); the diagonal corresponds to perfect calibration. The map that never forgets predicts 1 for every object (single point).}
  \label{fig:reliability}
\end{figure}
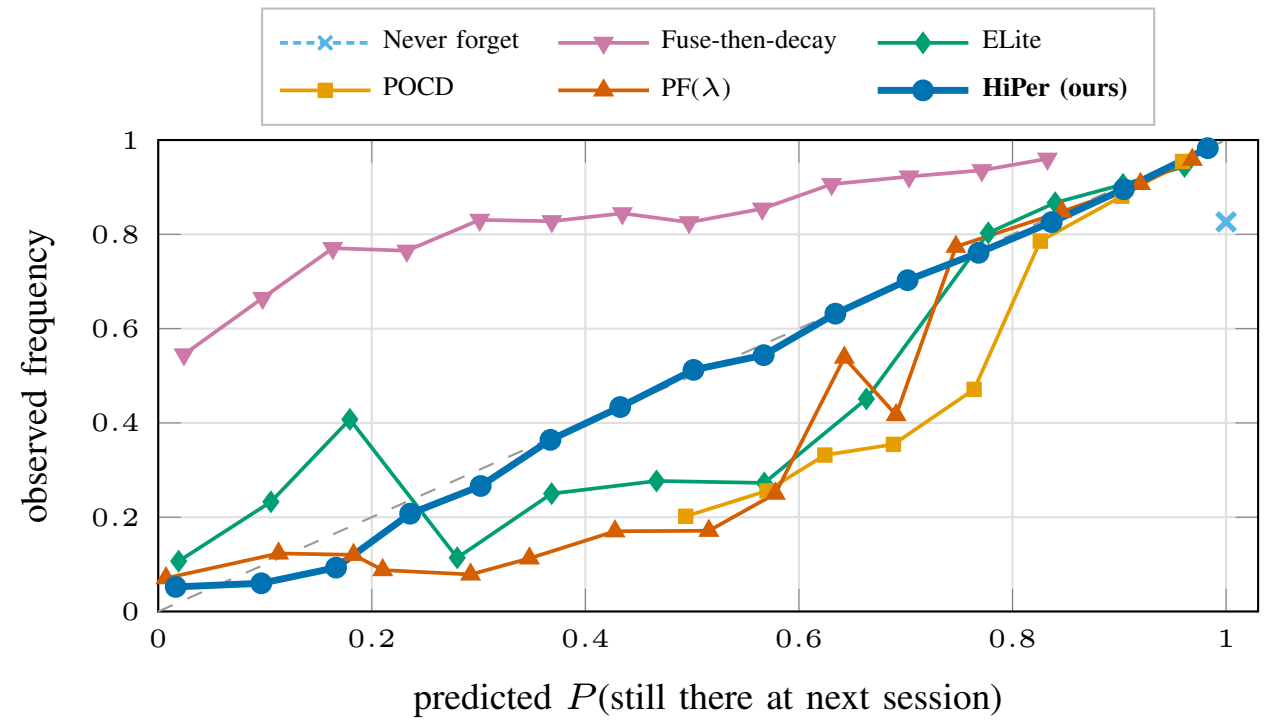

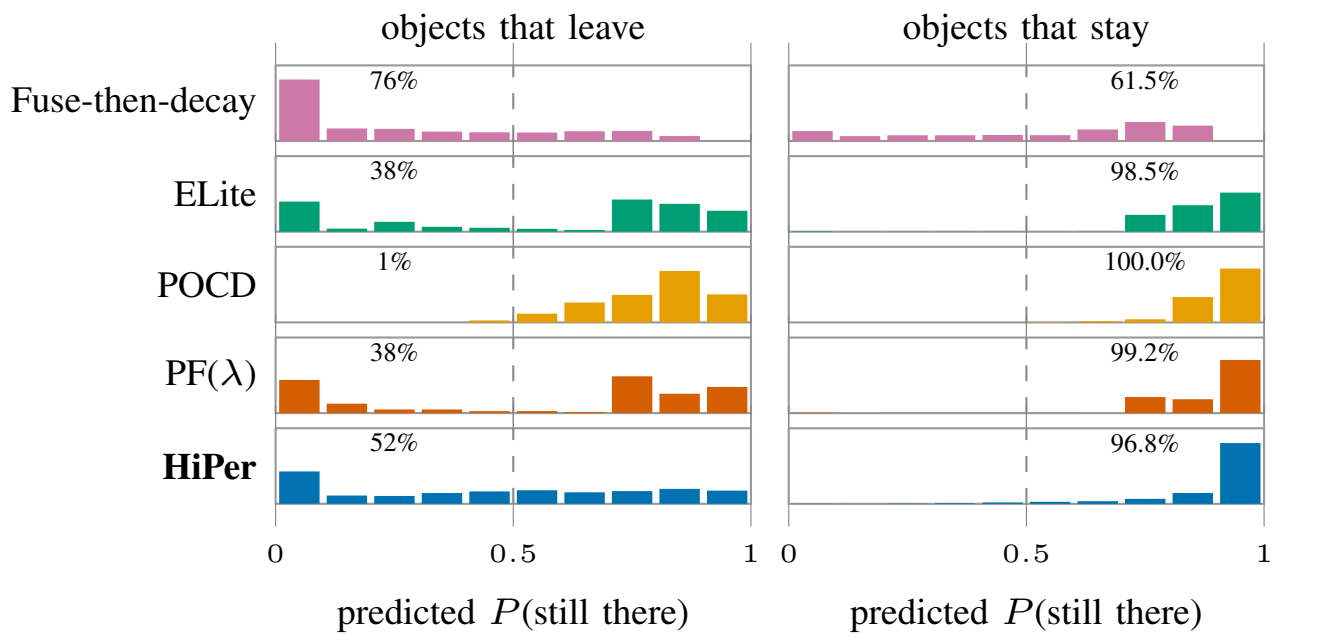
\begin{figure}[t]
  \centering
\begin{tikzpicture}
\begin{groupplot}[
    group style={group size=2 by 5, horizontal sep=0.25cm,
                 vertical sep=0.1cm, x descriptions at=edge bottom,
                 y descriptions at=edge left},
    scale only axis, width=0.355\columnwidth, height=0.5cm,
    xmin=0, xmax=1, ymin=0, xtick={0,0.5,1}, ytick=\empty,
    ybar, /pgf/bar width=0.085, /pgf/bar shift=0pt,
    axis line style={black!40}, tick label style={font=\tiny},
    label style={font=\scriptsize},
    xlabel={predicted $P$(still there)},
    ylabel style={rotate=-90, anchor=east, font=\scriptsize}]
  \nextgroupplot[ymax=55, ylabel={Fuse-then-decay}]
    \addplot[m ftd, mark=none, fill=., draw=none] table[x=mid, y expr=100*\thisrow{draft_v1_leave}, col sep=comma] {figures/data/pred_hist.csv};
    \draw[black!45, dashed, line width=0.4pt] (axis cs:0.5,0) -- (axis cs:0.5,55);
    \node[font=\tiny, anchor=north, inner sep=0.5pt] at (axis cs:0.25,54) {\xLeaveLowDraft\%};
  \nextgroupplot[ymax=90]
    \addplot[m ftd, mark=none, fill=., draw=none] table[x=mid, y expr=100*\thisrow{draft_v1_stay}, col sep=comma] {figures/data/pred_hist.csv};
    \draw[black!45, dashed, line width=0.4pt] (axis cs:0.5,0) -- (axis cs:0.5,90);
    \node[font=\tiny, anchor=north, inner sep=0.5pt] at (axis cs:0.75,88) {\xStayHighDraft\%};
  \nextgroupplot[ymax=55, ylabel={ELite}]
    \addplot[m elite, mark=none, fill=., draw=none] table[x=mid, y expr=100*\thisrow{elite_leave}, col sep=comma] {figures/data/pred_hist.csv};
    \draw[black!45, dashed, line width=0.4pt] (axis cs:0.5,0) -- (axis cs:0.5,55);
    \node[font=\tiny, anchor=north, inner sep=0.5pt] at (axis cs:0.25,54) {\xLeaveLowElite\%};
  \nextgroupplot[ymax=90]
    \addplot[m elite, mark=none, fill=., draw=none] table[x=mid, y expr=100*\thisrow{elite_stay}, col sep=comma] {figures/data/pred_hist.csv};
    \draw[black!45, dashed, line width=0.4pt] (axis cs:0.5,0) -- (axis cs:0.5,90);
    \node[font=\tiny, anchor=north, inner sep=0.5pt] at (axis cs:0.75,88) {\xStayHighElite\%};
  \nextgroupplot[ymax=55, ylabel={POCD}]
    \addplot[m pocd, mark=none, fill=., draw=none] table[x=mid, y expr=100*\thisrow{pocd_leave}, col sep=comma] {figures/data/pred_hist.csv};
    \draw[black!45, dashed, line width=0.4pt] (axis cs:0.5,0) -- (axis cs:0.5,55);
    \node[font=\tiny, anchor=north, inner sep=0.5pt] at (axis cs:0.25,54) {\xLeaveLowPocd\%};
  \nextgroupplot[ymax=90]
    \addplot[m pocd, mark=none, fill=., draw=none] table[x=mid, y expr=100*\thisrow{pocd_stay}, col sep=comma] {figures/data/pred_hist.csv};
    \draw[black!45, dashed, line width=0.4pt] (axis cs:0.5,0) -- (axis cs:0.5,90);
    \node[font=\tiny, anchor=north, inner sep=0.5pt] at (axis cs:0.75,88) {\xStayHighPocd\%};
  \nextgroupplot[ymax=55, ylabel={PF($\lambda$)}]
    \addplot[m pf, mark=none, fill=., draw=none] table[x=mid, y expr=100*\thisrow{pf_leave}, col sep=comma] {figures/data/pred_hist.csv};
    \draw[black!45, dashed, line width=0.4pt] (axis cs:0.5,0) -- (axis cs:0.5,55);
    \node[font=\tiny, anchor=north, inner sep=0.5pt] at (axis cs:0.25,54) {\xLeaveLowPf\%};
  \nextgroupplot[ymax=90]
    \addplot[m pf, mark=none, fill=., draw=none] table[x=mid, y expr=100*\thisrow{pf_stay}, col sep=comma] {figures/data/pred_hist.csv};
    \draw[black!45, dashed, line width=0.4pt] (axis cs:0.5,0) -- (axis cs:0.5,90);
    \node[font=\tiny, anchor=north, inner sep=0.5pt] at (axis cs:0.75,88) {\xStayHighPf\%};
  \nextgroupplot[ymax=55, ylabel={\textbf{HiPer}}]
    \addplot[m hiper, mark=none, fill=., draw=none] table[x=mid, y expr=100*\thisrow{ours_leave}, col sep=comma] {figures/data/pred_hist.csv};
    \draw[black!45, dashed, line width=0.4pt] (axis cs:0.5,0) -- (axis cs:0.5,55);
    \node[font=\tiny, anchor=north, inner sep=0.5pt] at (axis cs:0.25,54) {\xLeaveLowOurs\%};
  \nextgroupplot[ymax=90]
    \addplot[m hiper, mark=none, fill=., draw=none] table[x=mid, y expr=100*\thisrow{ours_stay}, col sep=comma] {figures/data/pred_hist.csv};
    \draw[black!45, dashed, line width=0.4pt] (axis cs:0.5,0) -- (axis cs:0.5,90);
    \node[font=\tiny, anchor=north, inner sep=0.5pt] at (axis cs:0.75,88) {\xStayHighOurs\%};
\end{groupplot}
\node[font=\scriptsize, anchor=south] at (group c1r1.north) {objects that leave};
\node[font=\scriptsize, anchor=south] at (group c2r1.north) {objects that stay};
\end{tikzpicture}
  \caption{Predicted presence for objects that leave before the next session (left) and that stay (right), all \xPredN{} held-out predictions. }
  \label{fig:predhist}
\end{figure}

Table~\ref{tab:persistence} summarizes the results on the held-out THOR-Lifelong sequences. \hiper achieves the lowest NLL (\rThorOursNll) and Brier score (\rThorOursBrier) and the highest AUROC (\rThorOursAuroc) of all compared methods. Every baseline, including PF($\lambda$) (\rThorPfNll), is significantly worse in all three metrics. Calibration lowers the NLL of every method and benefits the weaker baselines most (POCD improves from \rThorPocdNll{} to \rThorPocdCalNll). After calibration, \hiper still attains the lowest NLL (\rThorOursCalNll), and every baseline remains significantly worse in each of the three held-out sets. Among the re-tuned ablations, PF with the retrieved rates comes closest after calibration (\rThorPfnCalNll). This margin is significant over the pooled sequences, also when floor plans instead of sequences are resampled, but not within two of the three sets alone. Each level of the hierarchy contributes to the result (Sec.~\ref{sec:ablation}). The optional environment rate (Sec.~\ref{sec:env}; not shown in the table) slightly increases the NLL in this matched setting (\rThorEnvNll). PF($\lambda$) is significantly better calibrated, both raw (ECE \rThorPfEce{} vs.\ \rThorOursEce) and after calibration (\rThorPfCalEce{} vs.\ \rThorOursCalEce). However, it discriminates poorly (AUROC \rThorPfAuroc), since a single rate separates objects only by their detection histories (Fig.~\ref{fig:reliability}). The improvement is consistent across sequences: \hiper attains a lower NLL than PF($\lambda$) in \xWinNllPf{} of the \xSeqN{} sequences, and than ELite and POCD in \xWinNllElite{} and \xWinNllPocd. It also makes about half as many confident errors as PF($\lambda$) (\xConfErrOurs{} vs.\ \xConfErrPf; Fig.~\ref{fig:predhist}).

The hand-set table degrades the predictions at our parameters (NLL \rThorHandAblNll, Table~\ref{tab:ablation}). With ground-truth labels instead of CLIP labels, the table ranks objects well (AUROC \rThorHandGtAuroc) but remains miscalibrated (NLL \rThorHandGtNll). Its ordering is informative, but its lifetimes do not match those of the scenes. Since our simulator draws hazards by affordance rather than by class, a world whose changes follow object classes could favor hand-set tables more.

For map maintenance evaluation, \hiper and PF($\lambda$) prune similarly, with premature-prune rates of \rThorOursPrem{} and \rThorPfPrem{} and almost no stale objects. ELite's scalar belief has a premature-prune rate of \rThorElitePrem. The POCD belief never removes an object in our session-level replay (stale rate \rThorPocdStale), since its out-of-view decay stops at a floor and a single miss per session does not cross its removal threshold. The fuse-then-decay filter retains most departed objects (stale rate \rThorDraftStale).

\begin{figure}[t]
  \centering
\begin{tikzpicture}
\begin{groupplot}[
    group style={group size=2 by 1, horizontal sep=0.3cm,
                 y descriptions at=edge left},
    scale only axis, width=0.4\columnwidth, height=2.6cm,
    xmin=-0.3, xmax=6.3, ymin=0.21, ymax=0.45, xtick={0,...,6},
    ytick={0.25,0.3,0.35,0.4},
    yticklabel style={/pgf/number format/fixed, /pgf/number format/precision=2},
    xlabel={session $k$}, label style={font=\scriptsize},
    tick label style={font=\tiny},
    title style={font=\scriptsize, yshift=-1.4mm},
    grid=major, grid style={black!12}, axis line style={black!60},
    legend style={font=\tiny, draw=black!25, legend columns=3,
                  /tikz/every even column/.append style={column sep=5pt}},
    legend cell align=left]
  \nextgroupplot[title={raw}, ylabel={NLL (lower is better)},
                 legend to name=sessionlegend]
    \methodlegend
    \addplot[name path=pfline, draw=none, forget plot] table[x=k, y=pf, col sep=comma] {figures/data/nll_by_session.csv};
    \addplot[name path=ourline, draw=none, forget plot] table[x=k, y=ours, col sep=comma] {figures/data/nll_by_session.csv};
    \addplot[okBlue!12, forget plot] fill between[of=pfline and ourline];
    \addplot[name path=oh, draw=none, forget plot] table[x=k, y=ours_hi, col sep=comma] {figures/data/nll_by_session.csv};
    \addplot[name path=ol, draw=none, forget plot] table[x=k, y=ours_lo, col sep=comma] {figures/data/nll_by_session.csv};
    \addplot[okBlue!30, forget plot] fill between[of=oh and ol];
    \addplot[m elite, forget plot] table[x=k, y=elite, col sep=comma] {figures/data/nll_by_session.csv};
    \addplot[m pocd, forget plot] table[x=k, y=pocd, col sep=comma] {figures/data/nll_by_session.csv};
    \addplot[m pf, forget plot] table[x=k, y=pf, col sep=comma] {figures/data/nll_by_session.csv};
    \addplot[m hiper, forget plot] table[x=k, y=ours, col sep=comma] {figures/data/nll_by_session.csv};
  \nextgroupplot[title={after calibration}]
    \addplot[name path=ch, draw=none, forget plot] table[x=k, y=ours_cal_hi, col sep=comma] {figures/data/nll_by_session.csv};
    \addplot[name path=cl, draw=none, forget plot] table[x=k, y=ours_cal_lo, col sep=comma] {figures/data/nll_by_session.csv};
    \addplot[okBlue!30, forget plot] fill between[of=ch and cl];
    \addplot[m never, forget plot] table[x=k, y=always_cal, col sep=comma] {figures/data/nll_by_session.csv};
    \addplot[m ftd, forget plot] table[x=k, y=draft_v1_cal, col sep=comma] {figures/data/nll_by_session.csv};
    \addplot[m elite, forget plot] table[x=k, y=elite_cal, col sep=comma] {figures/data/nll_by_session.csv};
    \addplot[m pocd, forget plot] table[x=k, y=pocd_cal, col sep=comma] {figures/data/nll_by_session.csv};
    \addplot[m pf, forget plot] table[x=k, y=pf_cal, col sep=comma] {figures/data/nll_by_session.csv};
    \addplot[m hiper, forget plot] table[x=k, y=ours_cal, col sep=comma] {figures/data/nll_by_session.csv};
\end{groupplot}
\path (group c1r1.north east) -- node[above=13pt]
  {\pgfplotslegendfromname{sessionlegend}} (group c2r1.north west);
\end{tikzpicture}
  \caption{NLL per session on the held-out sequences, raw (left) and after calibration (right). Band: 95\% interval of \hiper; shaded: gain over PF($\lambda$). Raw fuse-then-decay (\xNllSessDraftMin{} to \xNllSessDraftMax) and never forget (\xNllSessNeverMin{} to \xNllSessNeverMax) lie above the range.}
  \label{fig:sessions}
\end{figure}
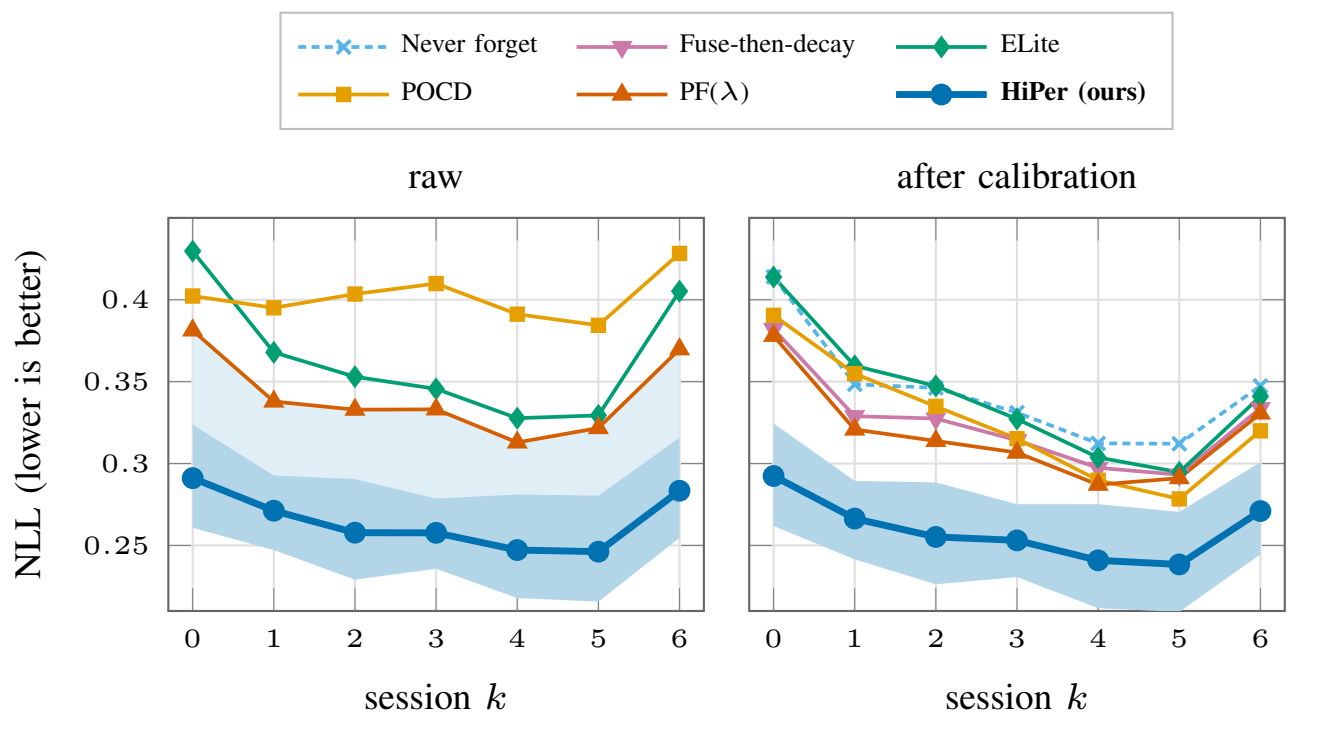

Fig.~\ref{fig:sessions} shows the NLL per session. \hiper remains below PF($\lambda$) after every session, by a margin of \xGapMin{} to \xGapMax. This holds from the prediction after session 0 (\xNllFirstOurs{} vs.\ \xNllFirstPf), when most objects are informed mainly by the retrieved prior, to the prediction after session 6 (\xNllLastOurs{} vs.\ \xNllLastPf). After calibration, \hiper has the lowest NLL of all methods after every session, by at least \xCalGapMin. Per object group (Table~\ref{tab:groups}), \hiper attains the lowest NLL in all three groups. For moveable objects, which change rarely, the margin is small and PF($\lambda$) ranks them slightly better (higher AUROC). Both learned priors improve with more training data (validation set, frozen parameters, one run each). Fitted on \xTsFewScenes, \xTsMidScenes{} and \xTsAllScenes{} training floor plans, the retrieved prior yields an NLL of \xTsFewOurs, \xTsMidOurs{} and \xTsAllOurs, and the CLIP class rates yield \xTsFewCls, \xTsMidCls{} and \xTsAllCls.

As a diagnostic, we provide the filter with the true simulated hazard of every object as its prior, tuned on validation over a grid comparable to those of the other methods. This oracle ranks objects slightly better (validation AUROC \xOracleAuroc{} vs.\ \xValOursAuroc; Brier score \xOracleBrier{} vs.\ \xValOursBrier) but does not achieve a lower NLL (\xOracleNll{} vs.\ \xValOursNll; \xOraclePfNll{} without per-object learning). Which object leaves next is largely random, and \hiper's learned prior also absorbs the noise of the mapping pipeline, which a true rate does not describe. Further gains in log-likelihood are therefore more likely to come from better evidence than from a better prior.

\begin{table}[t]
\caption{Results on the held-out THOR-Lifelong sequences per object group.}
\label{tab:groups}
\centering\scriptsize\setlength{\tabcolsep}{1.05pt}
\begin{tabular*}{\linewidth}{@{\extracolsep{\fill}}l|cc|cc|c@{}}
\toprule
 \multirow{2}{*}{Method}
 & \multicolumn{2}{c}{pickupable ($n$=16669)} & \multicolumn{2}{c}{moveable ($n$=9866)} & fixed ($n$=15839)\\
 & NLL$\downarrow$ & AUC$\uparrow$ & NLL$\downarrow$ & AUC$\uparrow$ & NLL$\downarrow$\\
\midrule
ELite scalar~\cite{gil2025elite} & 0.621\,{\tiny$\pm$0.111} & 0.728\,{\tiny$\pm$0.075} & 0.278\,{\tiny$\pm$0.176} & 0.502\,{\tiny$\pm$0.224} & 0.125\,{\tiny$\pm$0.129}\\
POCD~\cite{bogenberger2026glasses} & 0.748\,{\tiny$\pm$0.112} & 0.676\,{\tiny$\pm$0.078} & 0.279\,{\tiny$\pm$0.204} & 0.504\,{\tiny$\pm$0.222} & \underline{0.079}\,{\tiny$\pm$0.013}\\

PF($\lambda$)~\cite{rosen2016persistence} & \underline{0.563}\,{\tiny$\pm$0.110} & \underline{0.816}\,{\tiny$\pm$0.057} & \underline{0.243}\,{\tiny$\pm$0.155} & \textbf{0.722}\,{\tiny$\pm$0.160} & 0.149\,{\tiny$\pm$0.186}\\

HiPer & \textbf{0.481}\,{\tiny$\pm$0.079} & \textbf{0.828}\,{\tiny$\pm$0.052} & \textbf{0.237}\,{\tiny$\pm$0.182} & \underline{0.716}\,{\tiny$\pm$0.168} & \textbf{0.026}\,{\tiny$\pm$0.046}\\

\bottomrule
\end{tabular*}
\end{table}

\subsection{Real Data}
\label{sec:real}

\begin{table*}[t]
\caption{Results on the 47 3RScan test environments (mean $\pm$ standard deviation over environments). Calibrated: with calibration maps fitted on the tuning environments.}
\label{tab:rscan}
\centering\scriptsize\setlength{\tabcolsep}{3.0pt}
\begin{tabularx}{\linewidth}{@{}l||YYYY|YY|YY@{}}
\toprule
\multirow{2}{*}{Method}
 & \multicolumn{4}{c|}{Prediction} & \multicolumn{2}{c|}{Calibrated} & \multicolumn{2}{c}{Map maintenance}\\
 & NLL$\downarrow$ & Brier$\downarrow$ & ECE$\downarrow$ & AUC$\uparrow$ & NLL$\downarrow$ & ECE$\downarrow$ & Prem.$\downarrow$ & Stale$\downarrow$\\
\midrule
Never forget~\cite{cadena2016past} & 1.754\,{\tiny$\pm$1.857} & 0.127\,{\tiny$\pm$0.134} & 0.127\,{\tiny$\pm$0.134} & 0.500\,{\tiny$\pm$0.000} & 0.371\,{\tiny$\pm$0.277} & \underline{0.099}\,{\tiny$\pm$0.090} & -- & 1.000\,{\tiny$\pm$0.000}\\

Fuse-then-decay~\cite{deng2023ltvs} & 0.551\,{\tiny$\pm$0.076} & 0.181\,{\tiny$\pm$0.036} & 0.284\,{\tiny$\pm$0.092} & 0.622\,{\tiny$\pm$0.163} & 0.372\,{\tiny$\pm$0.280} & 0.108\,{\tiny$\pm$0.088} & 0.779\,{\tiny$\pm$0.373} & \underline{0.787}\,{\tiny$\pm$0.237}\\

ELite scalar~\cite{gil2025elite} & 0.382\,{\tiny$\pm$0.249} & 0.110\,{\tiny$\pm$0.095} & 0.117\,{\tiny$\pm$0.077} & 0.551\,{\tiny$\pm$0.124} & 0.371\,{\tiny$\pm$0.276} & 0.100\,{\tiny$\pm$0.086} & 0.667\,{\tiny$\pm$0.471} & 0.958\,{\tiny$\pm$0.102}\\

POCD~\cite{bogenberger2026glasses} & \underline{0.375}\,{\tiny$\pm$0.305} & \underline{0.108}\,{\tiny$\pm$0.106} & \underline{0.102}\,{\tiny$\pm$0.095} & \underline{0.650}\,{\tiny$\pm$0.158} & \underline{0.364}\,{\tiny$\pm$0.270} & 0.104\,{\tiny$\pm$0.084} & \underline{0.476}\,{\tiny$\pm$0.387} & \textbf{0.771}\,{\tiny$\pm$0.406}\\

PF($\lambda$)~\cite{rosen2016persistence} & 0.377\,{\tiny$\pm$0.298} & 0.109\,{\tiny$\pm$0.106} & \textbf{0.098}\,{\tiny$\pm$0.095} & 0.522\,{\tiny$\pm$0.127} & 0.369\,{\tiny$\pm$0.277} & \textbf{0.097}\,{\tiny$\pm$0.086} & 0.750\,{\tiny$\pm$0.354} & 0.958\,{\tiny$\pm$0.102}\\

HiPer (ours) & \textbf{0.310}\,{\tiny$\pm$0.228} & \textbf{0.093}\,{\tiny$\pm$0.084} & 0.122\,{\tiny$\pm$0.079} & \textbf{0.790}\,{\tiny$\pm$0.174} & \textbf{0.311}\,{\tiny$\pm$0.241} & 0.112\,{\tiny$\pm$0.080} & \textbf{0.467}\,{\tiny$\pm$0.506} & 0.792\,{\tiny$\pm$0.401}\\

\bottomrule
\end{tabularx}
\end{table*}

\begin{figure}[t]
  \centering
  \includegraphics[width=\columnwidth]{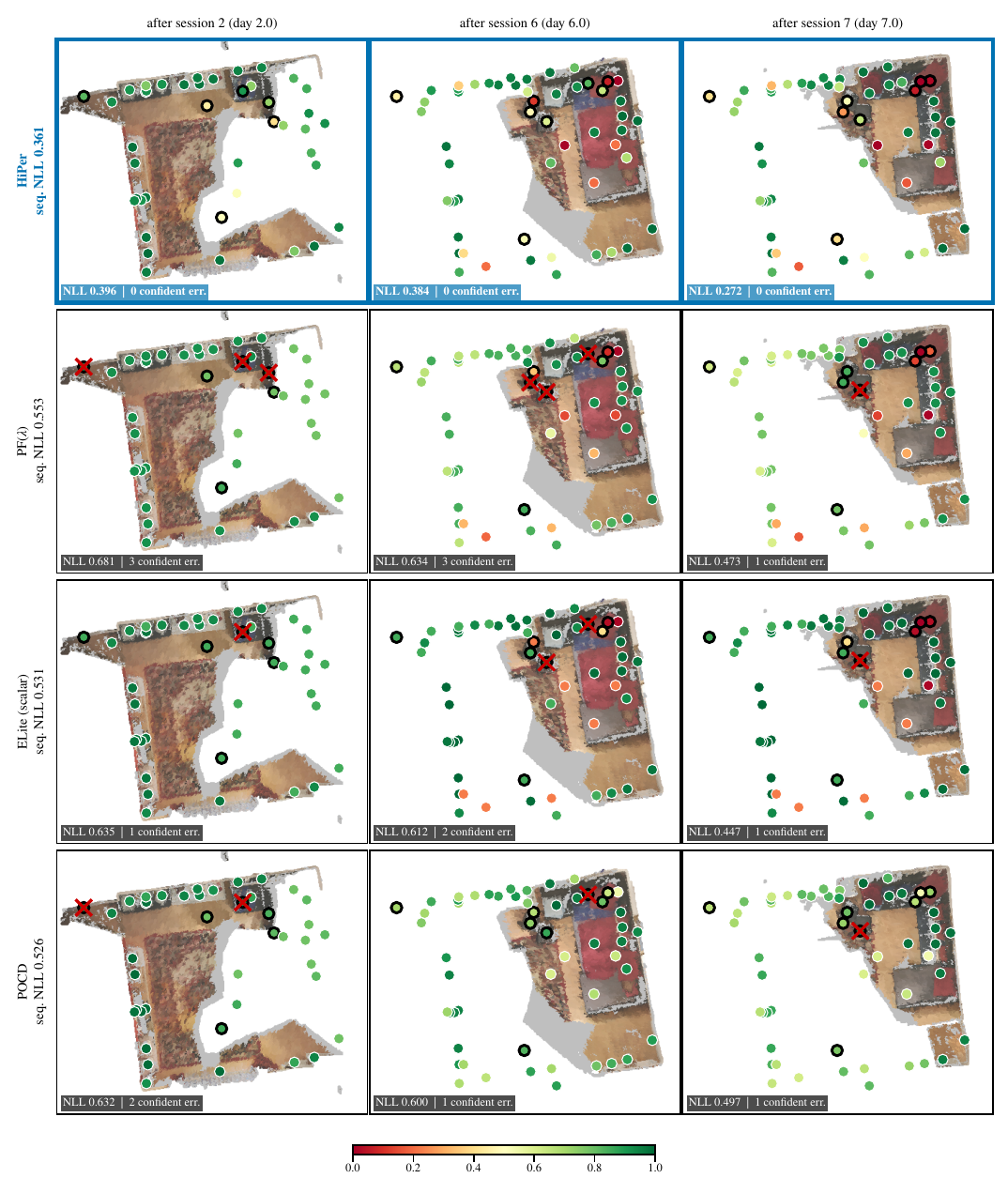}
  \caption{Qualitative results on the 3RScan environment (largest NLL margin over the baselines among environments with at least 50 labeled predictions). Each column shows the scan of one session from above (gray: unobserved) for the three sessions with the most departures. }
  \label{fig:qualrscan}
\end{figure}

Table~\ref{tab:rscan} reports the results on the \rRscanScenes{} 3RScan test environments (\rRscanLabeled{} labeled predictions). \hiper achieves the lowest NLL (\rRscanOursNll) and Brier score (\rRscanOursBrier) and the highest AUROC (\rRscanOursAuroc), and every baseline is significantly worse in these metrics and in NLL after calibration. As in simulation, PF($\lambda$) is better calibrated (ECE \rRscanPfEce{} vs.\ \rRscanOursEce) but ranks objects poorly (AUROC \rRscanPfAuroc). Map maintenance is hardly measurable on 3RScan: only \xRsDeparted{} objects leave, in \xRsDepartedEnvs{} environments, and only \xRsAbsentMiss{} of their \xRsAbsentObs{} absent observations are misses, since the later rescans mostly do not observe them. \hiper has the lowest premature-prune rate (\rRscanOursPrem) and a stale rate close to the lowest (\rRscanOursStale{} vs.\ \rRscanPocdStale), but no difference is significant. The gain stems from the retrieved prior, whereas per-object learning and relocation do not change the result on 3RScan (Sec.~\ref{sec:ablation}). Fig.~\ref{fig:qualrscan} shows \hiper's best environment.

\begin{table}[t]
\caption{Results on ObChange with leave-one-room-out tuning and cross-fitted priors, raw and after calibration (mean $\pm$ standard deviation over rooms). Bold: best; underlined: second best.}
\label{tab:obchange}
\centering\scriptsize\setlength{\tabcolsep}{0.6pt}
\begin{tabularx}{\linewidth}{@{}l|YYY|YY@{}}
\toprule
 & \multicolumn{3}{c|}{Leave-one-room-out} & \multicolumn{2}{c}{Calibrated}\\
Method & NLL$\downarrow$ & ECE$\downarrow$ & AUC$\uparrow$ & NLL$\downarrow$ & ECE$\downarrow$\\
\midrule
Never forget~\cite{cadena2016past} & 11.597\,{\tiny$\pm$2.350} & 0.839\,{\tiny$\pm$0.170} & 0.500\,{\tiny$\pm$0.000} & \textbf{0.445}\,{\tiny$\pm$0.266} & \textbf{0.156}\,{\tiny$\pm$0.092}\\

Fuse-then-decay~\cite{deng2023ltvs} & 0.594\,{\tiny$\pm$0.140} & 0.298\,{\tiny$\pm$0.058} & 0.594\,{\tiny$\pm$0.098} & 0.457\,{\tiny$\pm$0.278} & \underline{0.159}\,{\tiny$\pm$0.093}\\

ELite scalar~\cite{gil2025elite} & 0.545\,{\tiny$\pm$0.082} & 0.265\,{\tiny$\pm$0.096} & \underline{0.619}\,{\tiny$\pm$0.067} & \underline{0.446}\,{\tiny$\pm$0.267} & 0.160\,{\tiny$\pm$0.086}\\

POCD~\cite{bogenberger2026glasses} & 0.774\,{\tiny$\pm$0.054} & 0.400\,{\tiny$\pm$0.154} & 0.601\,{\tiny$\pm$0.160} & 0.464\,{\tiny$\pm$0.272} & 0.161\,{\tiny$\pm$0.093}\\

PF($\lambda$)~\cite{rosen2016persistence} & \textbf{0.427}\,{\tiny$\pm$0.318} & \textbf{0.108}\,{\tiny$\pm$0.128} & 0.544\,{\tiny$\pm$0.100} & \underline{0.446}\,{\tiny$\pm$0.266} & \textbf{0.156}\,{\tiny$\pm$0.093}\\

HiPer (ours) & \underline{0.498}\,{\tiny$\pm$0.270} & \underline{0.231}\,{\tiny$\pm$0.052} & \textbf{0.630}\,{\tiny$\pm$0.208} & 0.487\,{\tiny$\pm$0.306} & 0.194\,{\tiny$\pm$0.095}\\
\bottomrule
\end{tabularx}
\end{table}

\begin{figure}[t]
  \centering
  \includegraphics[width=\columnwidth]{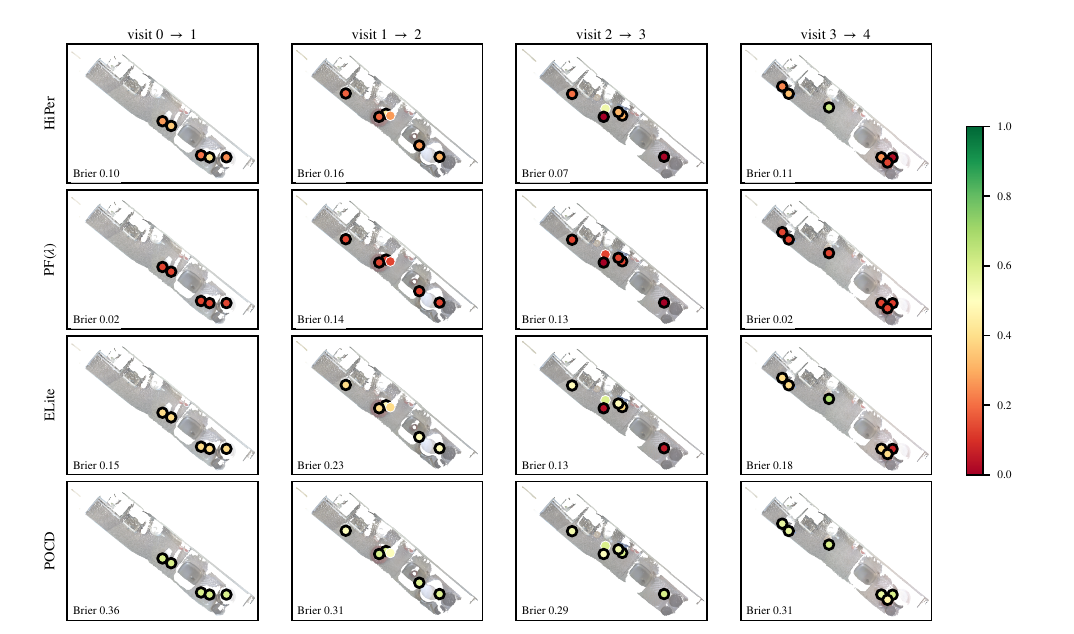}
  \caption{Qualitative results on the ObChange room. Objects on tables and shelves seen from above, colored by the predicted presence at the next visit; black ring: the object leaves. Each panel reports the Brier score.}
  \label{fig:qualreal}
\end{figure}

ObChange is a second, much smaller real dataset that rearranges objects between nearly every pair of visits. Models tuned in simulation, whose priors encode lifetimes of weeks, are therefore overconfident on it (\hiper: NLL \rObcOursNll, ECE \rObcOursEce; with the environment rate, \rObcEnvNll). After observing this transfer result, we added a leave-one-room-out evaluation (Table~\ref{tab:obchange}), in which the priors, parameters and calibration map of every method come from the other four rooms, with cross-fitted priors. PF($\lambda$) predicts best (NLL \rLoroPfNll) and outperforms \hiper (\rLoroOursNll) in \xLoroPfRoomsBetter{} of 5 rooms (exact sign-flip test, $p=\xLoroPfSignP$), also after calibration (\rLoroPfCalNll{} vs.\ \rLoroOursCalNll). Even the calibrated map that never forgets (\rLoroNeverCalNll) outperforms calibrated \hiper: with few objects and random rearrangements, the last evidence carries most of the signal (Fig.~\ref{fig:qualreal}). Since ObChange was analyzed in several rounds of development, it is weaker evidence than THOR and 3RScan.

\subsection{Ablation Study}
\label{sec:ablation}

\begin{table}[t]
\caption{Ablation study: \hiper with one component removed.All other rows keep \hiper's parameters.}\label{tab:ablation}
\centering\scriptsize\setlength{\tabcolsep}{1.0pt}
\begin{tabularx}{\linewidth}{@{}l|YY|YY@{}}
\toprule
\multirow{2}{*}{Variant}
 & \multicolumn{2}{c|}{THOR held-out} & \multicolumn{2}{c}{3RScan test}\\
 & NLL$\downarrow$ & AUC$\uparrow$ & NLL$\downarrow$ & AUC$\uparrow$\\
\midrule
\hiper (full) & \textbf{0.265}\,{\tiny$\pm$0.078} & \textbf{0.911}\,{\tiny$\pm$0.040} & \textbf{0.310}\,{\tiny$\pm$0.228} & \textbf{0.790}\,{\tiny$\pm$0.174}\\
\midrule
\multicolumn{5}{@{}l}{\emph{Prior}}\\
\hspace{0.5em}w/o retrieval: CLIP class rates & 0.296\,{\tiny$\pm$0.118} & 0.887\,{\tiny$\pm$0.040} & 0.322\,{\tiny$\pm$0.227} & 0.776\,{\tiny$\pm$0.187}\\
\hspace{0.5em}w/o retrieval: hand-set table & 0.545\,{\tiny$\pm$0.218} & 0.831\,{\tiny$\pm$0.062} & 0.521\,{\tiny$\pm$0.499} & 0.668\,{\tiny$\pm$0.159}\\
\hspace{0.5em}w/o CLIP: DINOv2-only key & 0.270\,{\tiny$\pm$0.080} & 0.908\,{\tiny$\pm$0.040} & 0.328\,{\tiny$\pm$0.238} & 0.733\,{\tiny$\pm$0.219}\\
\hspace{0.5em}w/o prior: one pooled rate & 0.326\,{\tiny$\pm$0.082} & 0.842\,{\tiny$\pm$0.044} & 0.369\,{\tiny$\pm$0.278} & 0.571\,{\tiny$\pm$0.151}\\
\midrule
\multicolumn{5}{@{}l}{\emph{W/o per-object belief}}\\
\hspace{0.5em}fixed rate (retrieved) & 0.274\,{\tiny$\pm$0.082} & 0.905\,{\tiny$\pm$0.041} & \underline{0.312}\,{\tiny$\pm$0.226} & \underline{0.787}\,{\tiny$\pm$0.162}\\
\hspace{0.5em}fixed rate (CLIP class) & 0.310\,{\tiny$\pm$0.107} & 0.869\,{\tiny$\pm$0.045} & 0.323\,{\tiny$\pm$0.222} & 0.778\,{\tiny$\pm$0.171}\\
\midrule
\multicolumn{5}{@{}l}{\emph{Relocation}}\\
\hspace{0.5em}w/o rate reset & 0.270\,{\tiny$\pm$0.084} & 0.908\,{\tiny$\pm$0.041} & 0.313\,{\tiny$\pm$0.233} & 0.785\,{\tiny$\pm$0.178}\\
\midrule
\multicolumn{5}{@{}l}{\emph{Association}}\\
\hspace{0.5em}w/o appearance & 0.394\,{\tiny$\pm$0.136} & 0.903\,{\tiny$\pm$0.035} & 0.334\,{\tiny$\pm$0.246} & 0.783\,{\tiny$\pm$0.167}\\
\bottomrule
\end{tabularx}
\end{table}

Table~\ref{tab:ablation} removes one component of \hiper at a time. \emph{Prior.} CLIP-label keys instead of retrieval raise the NLL on THOR from \rThorOursNll{} to \rThorClsNll{} and lower the AUROC from \rThorOursAuroc{} to \rThorClsAuroc; on 3RScan, they are significantly worse after calibration (\rRscanClsCalNll{} vs.\ \rRscanOursCalNll). A single pooled rate yields \rThorNoPriorNll{} on THOR and \rRscanNoPriorNll{} on 3RScan, and retrieval by DINOv2 alone yields \rThorDinoNll{} and \rRscanDinoNll; on 3RScan, both are significantly worse. \emph{Per-object belief and relocation.} On THOR, per-object learning lowers the NLL from \rThorPfnNll{} to \rThorOursNll, and resetting the rate at every relocation raises it to \rThorResetNll; both differences are significant. On 3RScan, neither component changes the result (\rRscanPfnNll{} and \rRscanResetNll), but \xRsKzeroPct\% of its predictions precede any history of the object, so the test has little power. Its \xRsLongEnvs{} environments with three or more rescans show no difference either (post hoc; NLL difference $\xRsLongDiff$, 95\% interval $\xRsLongLo$ to $\xRsLongHi$). \emph{Association.} \hiper re-identifies \xAsThorAll\% of the tracks of mapped objects on THOR (\xAsThorReloc\% of the relocated ones) and \xAsRsAll\% on 3RScan (\xAsRsReloc\%). Without appearance, few relocated objects are re-identified (\xAsThorGeoReloc\% and \xAsRsGeoReloc\%), and the NLL rises to \xAsThorGeoNll{} and \xAsRsGeoNll; without pose inflation, it rises slightly on THOR (\xAsThorNoPoseNll). Appearance also matches most novel objects to an existing one (\xAsThorNovel\% on THOR, \xAsRsNovel\% on 3RScan), partly because AI2-THOR renders every instance of a type with the same mesh. Since all beliefs replay the same evidence, their comparison is unaffected.

\subsection{Parameter Sensitivity and Runtime}
\label{sec:sensitivity}

\begin{table}[t]
\caption{Parameter sensitivity of \hiper on THOR-Lifelong validation (means over sequences): one parameter of the hierarchy or the prune threshold $\tau_\pi$ is changed, and the others keep their selected values (bold). $\tau_\pi$ changes only pruning; $\nu=0.1$ and $\tau_\pi=1$ lie outside the tuning grids.}
\label{tab:sensitivity}
\centering\scriptsize\setlength{\tabcolsep}{2pt}
\begin{tabularx}{\linewidth}{@{}c|c|YYYY|YY@{}}
\toprule
\multirow{2}{*}{Param.} & \multirow{2}{*}{Value}
 & \multicolumn{4}{c|}{Prediction} & \multicolumn{2}{c}{Map maint.}\\
 & & NLL$\downarrow$ & Brier$\downarrow$ & ECE$\downarrow$ & AUC$\uparrow$ & Prem.$\downarrow$ & Stale$\downarrow$\\
\midrule
\multirow{4}{*}{$\kappa_0$} & 8 & 0.285 & 0.086 & 0.054 & 0.898 & 0.096 & 0.019\\
 & 16 & 0.276 & 0.084 & 0.049 & 0.904 & 0.096 & 0.019\\
 & \textbf{32} & \textbf{0.274} & \textbf{0.083} & \textbf{0.050} & \textbf{0.906} & \textbf{0.085} & \textbf{0.019}\\
 & 64 & 0.275 & 0.084 & 0.052 & 0.906 & 0.080 & 0.013\\
\midrule
\multirow{4}{*}{$K$} & 10 & 0.277 & 0.084 & 0.051 & 0.905 & 0.080 & 0.013\\
 & \textbf{30} & \textbf{0.274} & \textbf{0.083} & \textbf{0.050} & \textbf{0.906} & \textbf{0.085} & \textbf{0.019}\\
 & 100 & 0.276 & 0.084 & 0.051 & 0.906 & 0.079 & 0.013\\
 & 300 & 0.278 & 0.084 & 0.052 & 0.905 & 0.080 & 0.013\\
\midrule
\multirow{4}{*}{$\nu$} & 0.1 & 0.274 & 0.083 & 0.051 & 0.906 & 0.084 & 0.019\\
 & \textbf{0.25} & \textbf{0.274} & \textbf{0.083} & \textbf{0.050} & \textbf{0.906} & \textbf{0.085} & \textbf{0.019}\\
 & 1 & 0.276 & 0.084 & 0.051 & 0.906 & 0.075 & 0.013\\
 & 4 & 0.285 & 0.085 & 0.053 & 0.901 & 0.075 & 0.013\\
\midrule
\multirow{4}{*}{$\tau_\pi$} & 0.2 &  &  &  &  & 0.045 & 0.137\\
 & 0.3 &  &  &  &  & 0.081 & 0.102\\
 & \textbf{0.5} &  &  &  &  & \textbf{0.085} & \textbf{0.019}\\
 & 1 &  &  &  &  & 0.955 & 0.000\\
\bottomrule
\end{tabularx}
\end{table}

Table~\ref{tab:sensitivity} replays the validation sequences with one parameter changed at a time. A weak prior ($\kappa_0=8$) and strong shrinkage toward the pooled rate ($\nu=4$) raise the NLL to \xSensKappaLow{} and \xSensNuHigh, compared with \xSensSelNll{} at the selected values. The other values of $\kappa_0$, $K$ and $\nu$ change the NLL by at most \xSensFlatMax, and so do the parameters not shown ($\tau$, $P_M$ and $P_F\le 0.05$; at most \xSensOtherMax). Only a high false-alarm rate degrades the NLL (\xSensPfHigh{} at $P_F=0.2$). The prune threshold trades stale objects for premature prunes: from $\tau_\pi=0.2$ to 1, the stale rate falls from \xSensStaleLow{} to zero, while the premature-prune rate rises from \xSensPremLow{} to \xSensPremHigh. The selected $\tau_\pi=0.5$ minimizes the mean of the two rates.

On a single thread of an Intel Core i7-13700KF CPU, association and evidence extraction take \dMapMs\,ms per session, and prior retrieval takes \dRetrieveMs\,ms per new object. A persistence-belief update requires only \dObserveUs\,$\mu$s, resulting in negligible overhead relative to perception.

\section{Conclusion}
\label{sec:conclusion}

We presented \hiper, a hierarchical persistence filter for lifelong object mapping. \hiper retrieves a survival prior from training objects with similar appearance and geometry, initializes an object-specific Bayesian persistence belief, and continuously adapts this belief from detections, misses, and relocations. By associating persistence with object identities rather than locations, the method learns how individual objects move over time. Experiments on \rThorScenes{} held-out simulated sequences demonstrated that \hiper achieves the best predictive performance among all compared methods in terms of NLL, Brier score, and AUROC, both with and without calibration. Retrieval-based priors consistently outperformed both zero-shot class-label priors and manually specified class-dependent lifetimes, while object-specific adaptation provided additional gains. Notably, \hiper matched or exceeded the performance of a filter given the true object-specific persistence rates. On held-out 3RScan environments, the retrieved prior again outperformed fixed-rate and published baselines, although the limited number of revisits provided insufficient evidence to realize the benefits of object-specific learning. In contrast, a small robot dataset with frequent random rearrangements favored a single globally tuned rate. These results highlight the conditions under which \hiper is most effective: environments containing objects similar to those seen during training, repeated observations over time, and object behaviors that exhibit temporal consistency. The current evaluation relies on oracle segmentation and tracking, and the available real-world datasets lack both reliable timestamps and sufficiently long observation histories for studying long-term persistence learning. Future work will integrate \hiper with learned perception and localization systems, evaluate it in extended real-world deployments, and explore richer persistence models, including learned mixture-based approaches such as Perpetua~\cite{saavedra2025perpetua}.

\bibliographystyle{IEEEtran} \bibliography{HiPer}

\end{document}